\documentclass[table]{article} % For LaTeX2e
\usepackage{iclr2025_conference,times}

\usepackage{amsmath,amsfonts,bm}

\def\eqref#1{equation~\ref{#1}}
\def\1{\bm{1}}

\DeclareMathAlphabet{\mathsfit}{\encodingdefault}{\sfdefault}{m}{sl}
\SetMathAlphabet{\mathsfit}{bold}{\encodingdefault}{\sfdefault}{bx}{n}

\usepackage{algorithm}
\usepackage{hyperref}
\usepackage{url}
\usepackage{natbib}
\usepackage{subfigure}
\usepackage{graphicx} 
\usepackage[utf8]{inputenc} % allow utf-8 input
\usepackage[T1]{fontenc}    % use 8-bit T1 fonts
\usepackage{hyperref}       % hyperlinks
\usepackage{url}            % simple URL typesetting
\usepackage{booktabs}       % professional-quality tables
\usepackage{amsfonts}       % blackboard math symbols
\usepackage{nicefrac}       % compact symbols for 1/2, etc.
\usepackage{microtype}      % microtypography
\usepackage{algorithmic}
\usepackage{multirow} 
\usepackage{graphicx}
\usepackage{subcaption}
\usepackage{enumitem}
\usepackage{dsfont}
\usepackage{titlesec}
\usepackage[font=small]{caption}
\usepackage[normalem]{ulem}
\newcommand{\jhc}[2]{\bgroup\textcolor{magenta}{\sout{#1} #2}\egroup}

\newcommand{\bstar}{\textsc{B-STaR}}

\newcommand{\bs}{\textsc{B-STaR}}

\title{\bstar: Monitoring and Balancing \\ Exploration and Exploitation in Self-Taught Reasoners}
\iclrfinalcopy
\author{Weihao Zeng\thanks{Equal Contribution.}\hspace{4pt}$^1$ \quad Yuzhen Huang$^{*1}$\quad Lulu Zhao$^2$\quad Yijun Wang$^3$\quad Zifei Shan$^3$ \quad Junxian He$^1$  \\
$^1$The Hong Kong University of Science and Technology \quad
$^2$BAAI
\quad 
$^3$Tencent \\
\texttt{\{wzengak,yhuanghj,junxianh\}@cse.ust.hk}
}

\begin{document}

\maketitle
\begin{abstract}
In the absence of extensive human-annotated data for complex reasoning tasks, self-improvement -- where models are trained on their own outputs -- has emerged as a primary method for enhancing performance. Recently, the approach to self-improvement has shifted toward a more dynamic, online fashion through iterative training. However, the critical factors underlying the mechanism of these iterative self-improving methods remain poorly understood, such as under what conditions self-improvement is effective, and what are the bottlenecks in the current iterations.
In this work, we identify and propose methods to monitor two pivotal factors in this iterative process: (1) the model's ability to generate sufficiently diverse responses (\emph{exploration}); and (2) the effectiveness of external rewards in distinguishing high-quality candidates from lower-quality
ones (\emph{exploitation}).
These factors are inherently dynamic throughout the iterative process, yet prior research rarely discusses their evolution -- leaving unclear why models often stagnate after only a few iterations.
Using mathematical reasoning as a case study, we begin with a quantitative analysis to track the dynamics of exploration and exploitation, discovering that a model's exploratory capabilities rapidly deteriorate over iterations, and the effectiveness of exploiting external rewards diminishes as well.
Motivated by these findings, we introduce \bs{}, a {\bf S}elf-{\bf Ta}ught {\bf R}easoning framework that autonomously adjusts configurations across iterations to {\bf B}alance exploration and exploitation, thereby optimizing the self-improving effectiveness based on the current policy model and available rewards.
Our experiments on mathematical reasoning, coding, and commonsense reasoning demonstrate that \bs{} not only enhances the model's exploratory capabilities throughout training but also achieves a more effective balance between exploration and exploitation, leading to superior performance. Crucially, this work deconstructs the opaque nature of self-training algorithms, providing interpretable insights into their dynamics and highlighting current limitations to guide future research.\footnote{We open-source our code at \href{https://github.com/hkust-nlp/B-STaR}{https://github.com/hkust-nlp/B-STaR}.}
\end{abstract}

\section{Introduction}

% Large language models possess advanced reasoning capabilities \citep{achiam2023gpt,dubey2024llama,liu2024deepseek} that allow them to generate step-by-step solutions \cite{wei2022chain} and identify errors when tackling complex tasks like mathematical problems \citep{cobbe2021training,hendrycks2021measuring} or coding challenges \citep{chen2021evaluating}. While training these models on extensive high-quality human-curated datasets can significantly enhance their reasoning abilities \citep{shao2024deepseekmath}, acquiring such data has become a major challenge in the field \citep{singh2023beyond}.

Large language models possess advanced reasoning capabilities such as mathematical problem-solving~\citep{cobbe2021training}, coding challenges~\citep{chen2021evaluating} or commonsense reasoning ~\citep{clark2018think}. However, 
the challenge of acquiring extensive, high-quality human-curated datasets remains a significant barrier to further enhancing these reasoning abilities. As tasks grow in complexity, the reliance on human-generated data becomes increasingly unsustainable, necessitating alternative approaches to training.

To tackle this issue, methods rooted in the concept of ``self-improvement''~\citep{huang2022large}, such as STaR~\citep{zelikman2022star}, RFT~\citep{yuan2023scaling}, and ReST~\citep{gulcehre2023reinforced,singh2023beyond}, provide more cost-effective and scalable solutions. Self-improvement follows an iterative process where the model generates responses, from which the better are selected to create higher-quality data for further refinement~\citep{hosseini2024v}. This continuous loop allows the model to improve its performance over time, reducing the need for large amounts of human-generated data. Ultimately, this approach enhances the model's ability to handle complex reasoning tasks, pushing the limits of its capabilities~\citep{havrilla2024teaching}.
 
% \begin{figure*}[t]
%     \centering
%     \subfigure[GSM8K]{
%         \includegraphics[width=0.23\textwidth]{fig/fig1_gsm8k.pdf}
%     }
%     \hfill
%     \subfigure[MATH]{
%         \includegraphics[width=0.23\textwidth]{fig/fig1_math.pdf}
%     }
%     \hfill
%     \subfigure[APPS]{
%         \includegraphics[width=0.23\textwidth]{fig/fig1_apps_v2.pdf}
%     }
%     \subfigure[ARC-C]{
%         \includegraphics[width=0.23\textwidth]{fig/fig1_arc.pdf}
%     }
%      \vspace{-5pt}
%     \caption{Pass@1 accuracy over training steps on GSM8K, MATH, APPS, ARC-Challenge. We compare multiple baselines, including SFT, Online RFT, STaR/ReST-EM, and Iterative RFT, against our proposed \bstar. For ARC-Challenge, we start from the Mistral-7b-instruct model, denoted as ``Base'' in the figure.}
%     \label{fig:main_pass@1}    
% \vspace{-35pt}
% \end{figure*}
Despite significant advancements, we still lack a deep understanding of the key factors that drive successful self-improvement and the internal optimization mechanisms remain largely opaque. Understanding the critical components and bottlenecks of these self-improving methods is particularly important given that performance of current self-improving approaches does not scale well with increased compute and saturates very quickly after merely 3 to 5 iterations~\citep{singh2023beyond, wu2024progress}. 
% Additionally, it remains unclear whether the gains in model capabilities during the self-improvement process are merely reflected in surface-level evaluation metrics or if they indicate more profound progress \cite{wu2024progress,wang2024planning}. 
In this paper, we address the following questions: (1) What key factors play a decisive role in the self-improvement process? (2) How can these factors be used to analyze the limitations of current self-improvement methods from a unified perspective? and (3) How can we leverage these factors to guide the self-improvement process, ultimately maximizing performance gains?

\begin{figure*}[t]
    \centering
    \subfigure[GSM8K]{
        \includegraphics[width=0.23\textwidth]{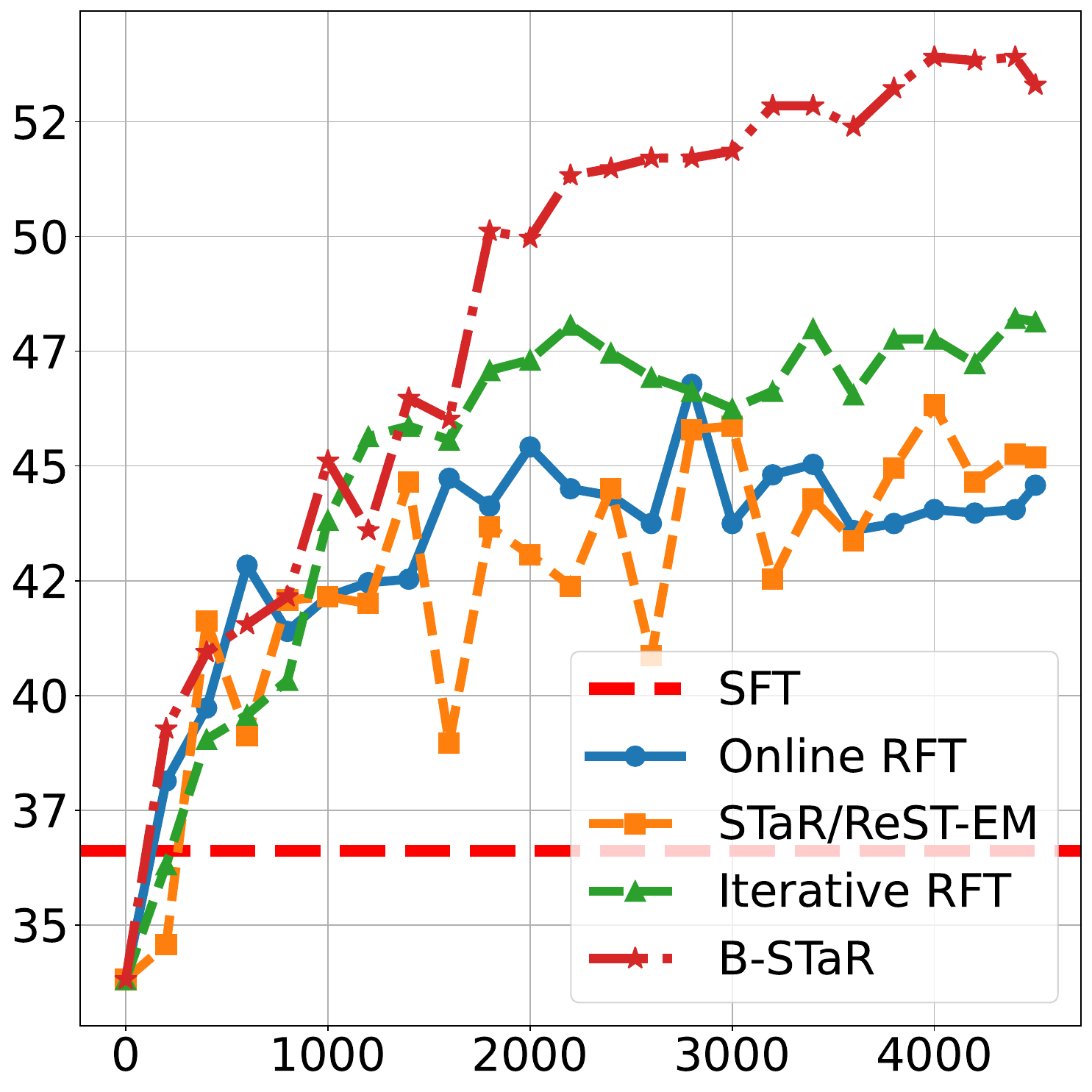}
    }
    \hfill
    \subfigure[MATH]{
        \includegraphics[width=0.23\textwidth]{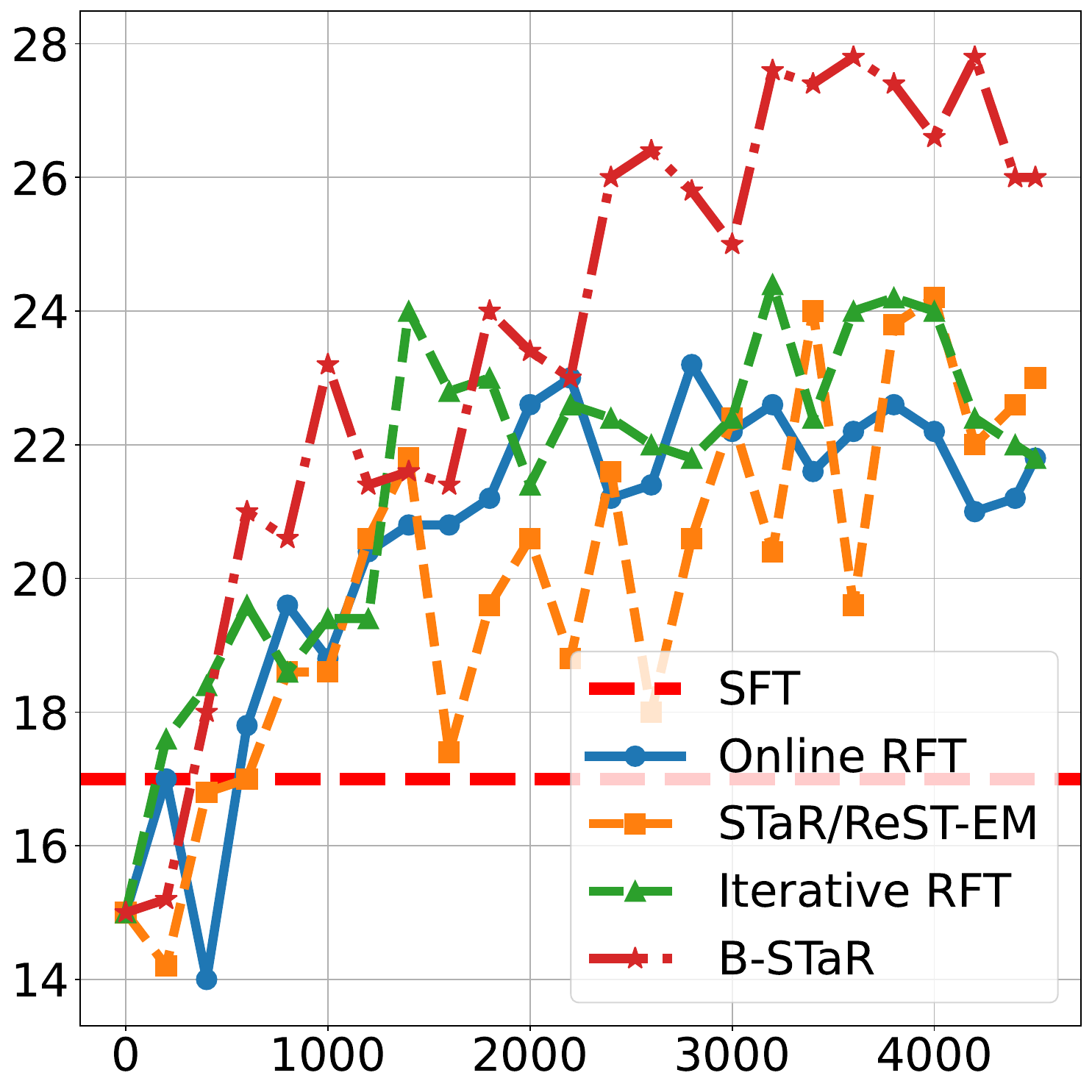}
    }
    \hfill
    \subfigure[APPS]{
        \includegraphics[width=0.23\textwidth]{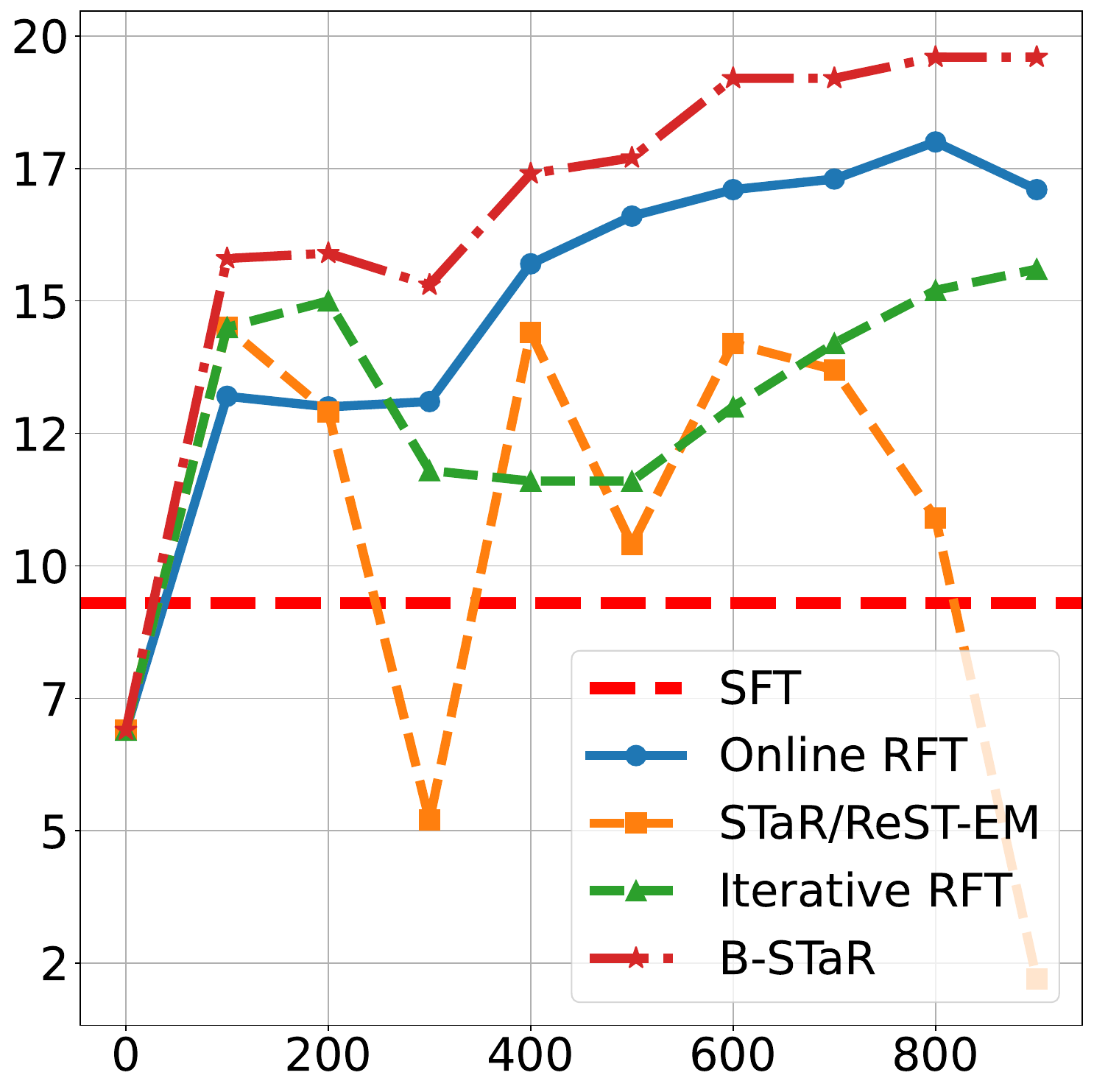}
    }
    \subfigure[ARC-C]{
        \includegraphics[width=0.23\textwidth]{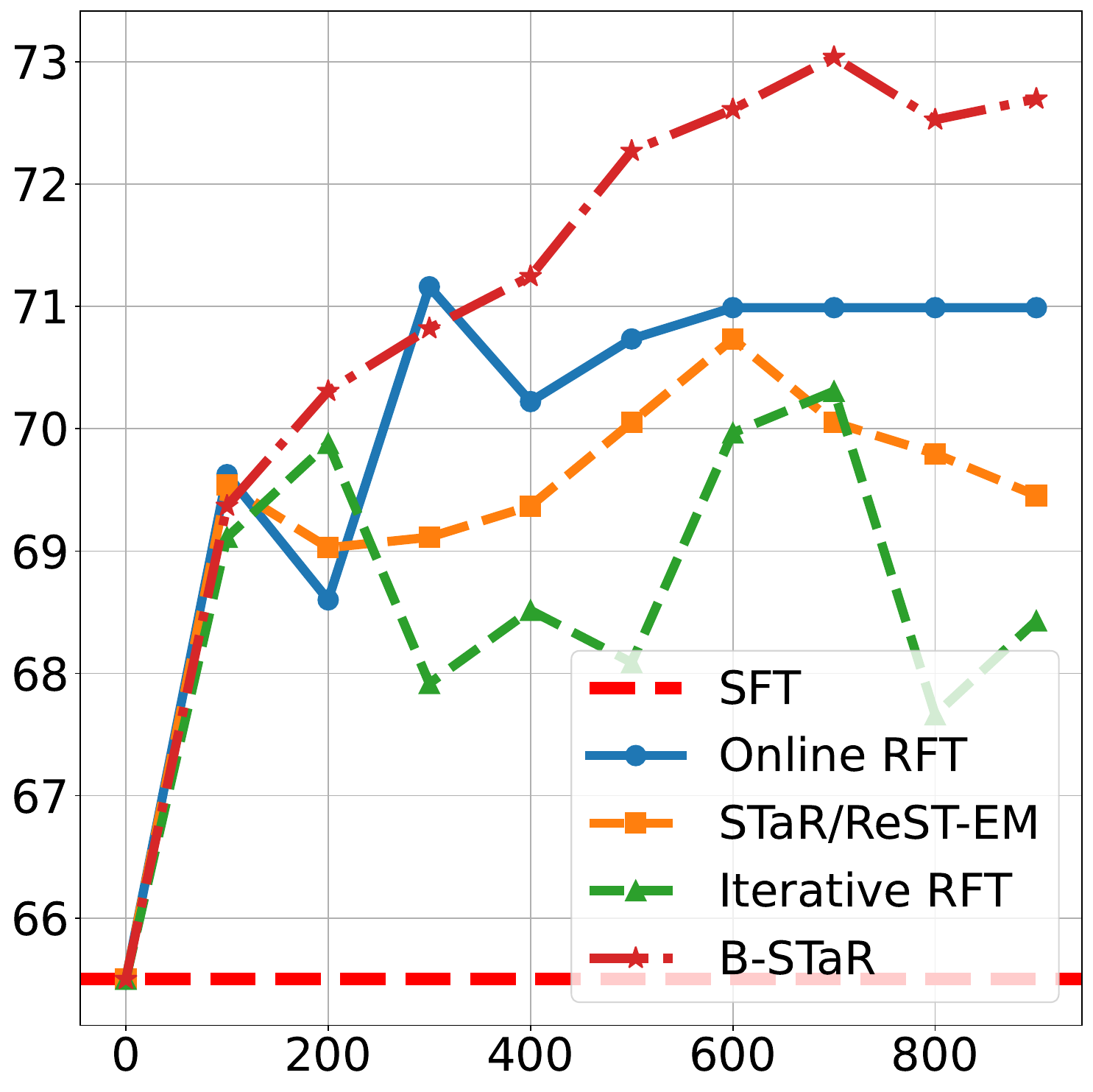}
    }
     \vspace{-5pt}
    \caption{Pass@1 accuracy over training steps on GSM8K, MATH, APPS, and ARC-Challenge. The GSM8K, MATH, and ARC-C results are from Mistral-7B, while the APPS scores are obtained based on Llama3-8B. We compare multiple baselines, including SFT, Online RFT, STaR/ReST-EM, and Iterative RFT, against our proposed \bstar. For further details on the experimental settings, please refer to \textsection\ref{sec:bstar_setup}.}
    \label{fig:main_pass@1}    
% \vspace{-40pt}
\end{figure*}

% \yh{I insert a new command here to fix the gap between text and figure. This command make the bottom flexible but might apply to all the rest of the paper.  If we decide to use it, plz carefully check. }
\raggedbottom

To this end, we identify two crucial capabilities of the model during its self-improvement process: (1) Exploration -- the model's ability to generate correct and diverse responses among multiple generated candidates~\citep{singh2023beyond,wang2024planning}, and (2) Exploitation -- the effectiveness of external rewards (e.g., a reward model or final answer supervision) in selecting high-quality solutions from these candidates~\citep{wang2024math,sun2024easy}.  
% \jhc{Connecting with traditional RL terminology, we correspond the two capabilities to \emph{exploration} and \emph{exploitation} respectively.}{} 
Intuitively, these two factors are dynamic, evolving throughout the training process, a phenomenon that remains underexplored despite its critical importance.
In this work, we first conduct empirical analysis to quantitatively monitor the dynamics of exploration and exploitation during iterative training processes. We observe that both capabilities may stagnate or even decline, and imbalances between them can hinder the model's ongoing improvement. 
% For instance, if the model produces nearly identical candidates for a given query when sampling multiple times, indicating weak exploration, then the impact of even the strongest reward is diminished in that training iteration. 

Motivated by these insights, we propose a novel approach for self-improvement that automatically monitors and balances these dynamic factors to optimize the use of the current policy and reward. This involves adjusting configurations that influence exploration and exploitation, such as sampling temperature and reward thresholds. 
These configurations are adaptively modified throughout training in terms of our proposed metric, \emph{balance score}.
This new metric assesses the potential of a query based on the current model's exploration and exploitation capabilities, and our method automatically balances exploration and exploitation behaviors to maximize the average balance scores. We refer to this method as \bstar, a Balanced Self-Taught Reasoner.
% given any query, which considers both exploration and exploitation and re unify these two abilities by considering them from the perspective of post-training data, as both ultimately aim to maximize the effectiveness of all queries during the self-improvement process. Our goal is to continuously enhance the positive impact of the data gathered in each iteration of optimization. To this end, we introduce a new metric, "Query Effect" to better capture these dynamics and guide the optimization process.

%To do this, we also introduce a new metric to measure this effect.

% By adopting this unified perspective, we can directly examine factors that influence Query Effect, such as Generating-related factors (e.g., the number of samples and the sampling temperature) and Rewarding-related factors (e.g., reward score strategies and the frequency of reward model updates). Additionally, we can leverage the Query Effect to dynamically adjust these influencing factors within a specified compute budget \cite{bansal2024smaller,snell2024scaling}, enabling sustained improvement in self-improvement performance.
% \jh{we need to talk about experimental results in the end of intro}

% \wz{include some math experimental results here}

The experimental results from mathematical problem-solving, coding challenges and commonsense reasoning demonstrate that \bstar{} significantly surpasses other self-improvement methods through balanced exploration and exploitation.
% in enhancing the model's exploration abilities, resulting in more robust and consistent performance improvements. 
For instance, \bstar{} achieves a significant improvement in Pass@1 on both GSM8K and MATH, surpassing various self-improving variants while maintaining a steady upward trajectory, as depicted in Figure \ref{fig:main_pass@1}. Furthermore, 
% as Figure \ref{fig:main_pass@k-s} illustrates, 
we demonstrate that exploration-related metrics, such as Pass@32, are continuously improving without any notable degradation. 
% which implies \bstar{} successfully balances exploration ability during training and overcomes the exploration issue. 
% This enhanced exploration capability enables the reward model to provide more effective supervision by consistently selecting high-quality solutions, as evidenced in Figure \ref{fig:main_reward@k-s}.

\begin{figure}[!t]
        \centering
        \includegraphics[width=\columnwidth]{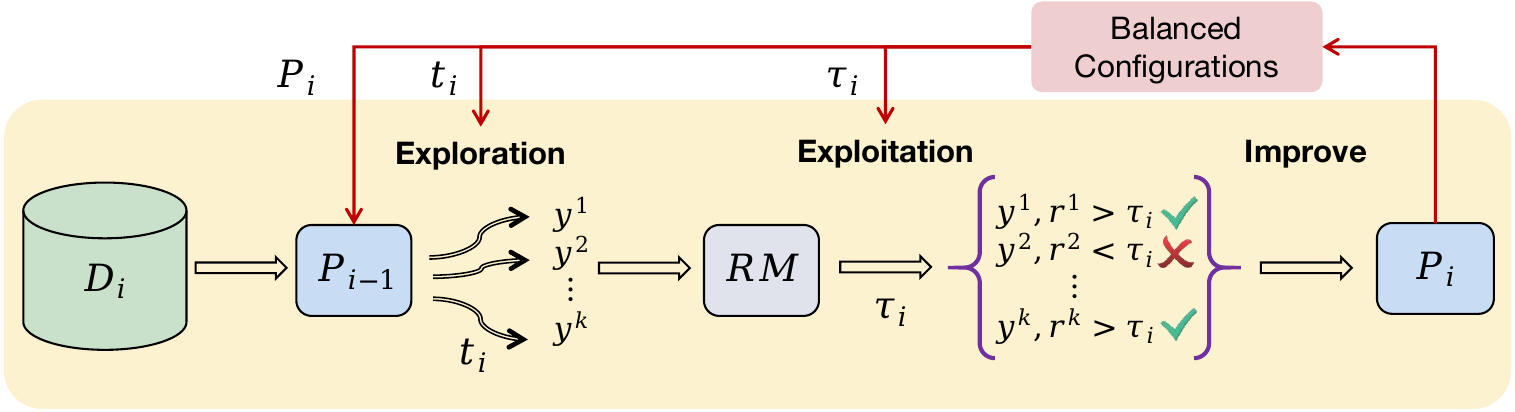}
        \caption{Illustration of the \bs{} approach. In each iteration, we first identify the configurations -- temperature $t_i$ and reward threshold $\tau_i$ -- that maximize the average balance scores using a small subset of training queries. Next, we apply the optimal temperature and threshold to generate and reward the full set of training queries. Finally, we update the model based on the selected data.
        }
        \label{fig:main}
    \vspace{-10pt}
\end{figure}

% \vspace{-30pt}
% \vspace{-20pt}
\section{Monitoring Exploration and Exploitation in Self-Improvement}

\label{sec:ana_self_improve}
% This section provides an overview of self-improvement, followed by an examination of how current self-improvement methods perform throughout the optimization process.

% This section starts with a brief overview of self-improvement, then we identify the critical factors in this process and propose methods to monitor them quantitatively. 
% followed by a quantitative analysis of the dynamic changes in exploration and exploitation throughout the training process.

\subsection{Background: Self-Improvement}
\label{sec:background}
% For the pre-trained model $M_{0}$ and a given training set $D={(x_{i},y_{i})}_{i=1}^{N}$ (where $x_{i}$ represents training queries and $y_{i}$ represents their correct solutions), the goal of self-improvement is to iteratively generate and select higher-quality solutions, followed by continuous refinement of the model's performance on the updated data. 

Given a pre-trained model $P_0$ and a training set $D = {(x_i, y_i)}_{i=1}^{N}$, where $x_i$ denotes the training query and $y_i$ represents the response,
% \jh{We are introducing the context and our method not specifically for problem solving, thus ``response'' is more general}, 
the goal of self-improvement is to iteratively generate high-quality responses from the current model and update the model with these self-generated data.
Let $T$ represent the total number of iterations, with the model at the start of the $t$-th iteration denoted as $P_{t-1}$. 
In the first iteration, $P_0$ is typically fine-tuned on the initial dataset $D$, then each subsequent iteration involves three critical steps generally ~\citep{yuan2023scaling, gulcehre2023reinforced}:

(1) Generating (Sampling): For each query $x_i$, the model $P_{t-1}$ generates $K$ candidate responses, forming a new, self-generated dataset.

(2) Rewarding (Verifying): A reward function $r(x, y)$ is applied to score and select the high-quality responses from the self-generated dataset. 
This reward can be binary and utilize additional supervision, for example, in problem-solving tasks in the math and code domains, it is common to use the final answer matching or the result of passing unit tests as binary feedback~\citep{yuan2023scaling,chen2022codet}.
In a more sophisticated case, $r(x,y)$ can be parameterized by a reward model outputting continuous scores, using outcome-based reward models (ORMs) ~\citep{li2022making} or process-based reward models (PRMs)~\citep{uesato2022solving,lightman2023let,havrilla2024teaching}.

% (2) Rewarding (Verifying): A reward mechanism selects from the self-generated dataset. Rewards can be sparse (e.g., +1 if the final answer is correct) or derived from reward models, such as outcome-based reward Models (ORMs) and process-based Reward Models (PRMs) \cite{uesato2022solving,lightman2023let,havrilla2024teaching}.

(3) Improving: The selected dataset is used to update $P_{t-1}$, producing $P_t$. 
To differentiate the generation model $M$ from the reward model, $M$ is also referred to as the policy model following RL literature.
% \jh{please use citet and citep correctly, as in our deita paper. This issue is in the entire paper as currently `cite` is used, please correct} 
In reasoning tasks that we are going to focus on, SFT loss is commonly used in the Improving step due to its robustness and scalability~\citep{pang2024iterative,dubey2024llama}, as more sophisticated RL losses can be unstable to optimize and scale up.
% Common training methods at this stage include Supervised Fine-Tuning (SFT) and Direct Preference Optimization (DPO)~\citep{rafailov2024direct}.
When SFT loss is adopted, the Rewarding step aims to reject low-quality responses and use remaining high-quality data for training, thus this process is also referred to as rejection fine-tuning (RFT,~\citet{yuan2023scaling}). In Appendix \ref{sec:concept_self}, we provide detailed explanations of self-improvement, including reward functions and RFT.

\paragraph{Discussion on Online Learning.}
% The iterative procedure described above can be contextualized within the reinforcement learning framework~\citep{singh2023beyond}, and the iterative design shifts the vanilla offline training towards a more dynamic online variant -- when iteration intervals are short and the optimizer is inherited between iterations, the training essentially transforms into a fully online learning algorithm.
% one iteration is offline training and the process is made iterative to update the model with the new on-policy data. 

% \yh{I think the original beginning is better in flow.}
% The iterative procedure described above can be contextualized within the reinforcement learning framework~\citep{singh2023beyond}. The iterative design shifts the vanilla offline training towards a more dynamic online variant. When iteration intervals are short and the optimizer is inherited between iterations, the training essentially transforms into a fully online learning algorithm. 

The iterative procedure described above can be contextualized within the reinforcement learning framework~\citep{singh2023beyond}, and the iterative design shifts the vanilla offline training towards a more dynamic online variant -- when iteration intervals are short and the optimizer is inherited between iterations, the training essentially transforms into a fully online learning algorithm.
% \jhc{For example, PPO~\citep{schulman2017proximal}, a prevalent online RL algorithm, exemplifies iterative training with small iteration intervals.}{}
% For example, PPO~\citep{schulman2017proximal}, a prevalent online RL algorithm, exemplifies iterative training with small iteration intervals.
% In problem-solving tasks that we are going to focus on in this paper, SFT loss is the most commonly used one in the Improving step due to its robustness and scalability\jh{cite several papers, including llama3}, as more sophisticated RL losses can be unstable to optimize and scale up. 
Iterative RFT implementations in previous works 
% STaR~\citep{zelikman2022star} and ReST\textsuperscript{EM}~\citep{singh2023beyond}, 
typically adopt long iteration intervals where each iteration processes all available queries~\citep{zelikman2022star,sun2024easy}. Sometimes, these implementations opted to restart the training from the initial checkpoint rather than from the last saved model~\citep{zelikman2022star,singh2023beyond}. However, we argue that always starting from the beginning is not scalable for large datasets, particularly in a streaming setting, instead, a continuous training approach more aligned with RL principles is preferable. Transitioning from offline to online training, online RFT~\citep{shao2024deepseekmath} has demonstrated its superiority over traditional offline RFT methods by switching iterations more frequently, ensuring that the synthetic data remains on-policy.
% Transitioning from offline to online training, Online RFT~\citep{shao2024deepseekmath} has demonstrated its superiority compared to traditional offline RFT methods. Compared to conventional iterative RFT, Online RFT switches iterations more frequently so that the synthetic data is always on-policy. 
In this study, we explore online RFT as our primary framework. 

\subsection{The Critical Factors -- Exploration and Exploitation}
\label{Sec:factor}
% Recently, the OpenAI o1 model leads to increasing interests in self-improvement, 
% To maximize the gains from self-improvement training and even lift model's ability fundamentally from its own outputs, the key is to achieve scalable self-improvement training, where the model performance is able to scale up with increased compute invested into the training algorithm.

To maximize the gains from self-improvement training and fundamentally enhance the model's ability using its own outputs, the key is to achieve scalable self-improvement training, where the model's performance scales with increased computational investment in the training algorithm.
However, all previous works show quick saturation after merely 3-5 self-improvement iterations~\citep{singh2023beyond, wu2024progress}, hypothesizing that the model's own outputs can only lead to limited gains.
In this work, we seek to dive deeper into the currently opaque process of self-improvement, to understand the critical factors that determine whether self-improvement succeeds or fails.

%make self-improvement successful or failed.  

% Intuitively, for a certain iteration of training, we argue that two high-level conditions must be met for the model to make progresses:
% (1) When multiple candidates are sampled from the model, they must include at least a small portion of high-quality responses. This is particularly important for the queries which the model fails to produce satisfied responses with greedy decoding, as models potentially learn the most from the queries which it fails to answer correctly. Such a condition requires the model to be able to explore diversely to yield good responses that are otherwise unreachable with greedy decoding; 
% (2) The reward function $r(x,y)$ in the rewarding step must be able to distinguish the good candidates from the bad ones reasonably well.
% If either of these two conditions are broken -- for example, when the model can only generate similar responses to the greedy-decoded ones, or the reward function cannot select the high-quality responses, the self-improvement will be limited on the gains if any. 

Intuitively, for a certain iteration of training, we argue that two high-level conditions must be met for the model to make progresses: (1) \emph{Diverse Exploration for High-Quality Responses:} When multiple candidates are sampled from the model, a portion of them must be high-quality responses. This is particularly important for queries where the model fails to produce satisfactory outputs using greedy decoding. 
Such diversity enables the discovery of responses that cannot be reached through greedy decoding. (2) \emph{Effective Reward Function Discrimination:} The reward function $r(x,y)$ must reliably distinguish high-quality candidates from lower-quality ones. 
We refer to the two conditions as \emph{exploration} and \emph{exploitation} respectively, and we provide their analogies to RL in Appendix~\ref{appendix:just-exp}.
If either of these conditions is unmet -- such as when the model produces responses overly similar (i.e. lack of diversity), or when the reward function fails to identify high-quality responses -- the self-improvement will be limited on the gains.

\paragraph{Exploration and Exploitation are Moving Targets.}
% \jh{introduce that these two are dynamically changing during training, and why there is a tradeoff, like if the model explores too diverse, that could be hard for the RM. There could be also distribution match as the RM if more off policy}

Both exploration and exploitation are dynamically influenced by the policy model during the self-improvement process. On one hand, after multiple iterations, the policy model may overfit the task, failing to explore diverse responses and instead generating highly similar outputs (i.e., a lack of exploration). Training on these highly similar responses is unlikely to yield significant improvements. On the other hand, if the model generates excessively diverse responses, resulting in a distribution that deviates significantly from the training data distribution of the reward model, it becomes challenging for the reward model to reliably distinguish high-quality responses (i.e., a lack of exploitation). Thus, maintaining a dynamic balance between exploration and exploitation is essential throughout the self-improving process. 
However, such dynamics are rarely discussed in prior research, 
% \jhc{while~\citet{wu2024progress} reveal a model's generation diversity tends to decline over the course of self-improvement, indicating a decrease in exploration capabilities.}{} 
% \jhc{We provide further justification of both exploration and exploitation in the context of RL in Appendix~\ref{appendix:just-exp}.}{} 
Next, we propose methods to quantify exploration and exploitation, enabling us to monitor their dynamics during training and deepen our understanding of the mechanisms underlying self-improvement.

\paragraph{Quantifying Exploration and Exploitation.}
% \jh{use bullet points to introduce the metrics for exploration and exploitation here. I guess some of the parts later can be moved here.}
In this work, we mainly focus on complex problem-solving tasks such as math and coding domains, where the correctness of the responses can be easily verified on labeled datasets.\footnote{Strictly speaking, the response may contain incorrect steps even though the final answer is correct, we do not further distinguish this difference following others~\citep{zelikman2022star} as it is not the focus of this work.}
 This property facilitates the quantification of exploration and exploitation, for which we detail the metrics below:
\begin{itemize}[leftmargin=*]
    \item \emph{Exploration:} Pass@K, which measures whether there is at least one correct response during K sampled candidates, is a straightforward metric for assessing exploration, as it directly reflects whether the model is able to explore correct solutions. 
    However, Pass@K can be noisy, as it only counts a single correct response, while it is desirable to assess whether the model can explore multiple correct response as well. 
    To this end, we propose to track Pass@K-S as well, which measures whether there are at least S unique correct responses among K sampled candidates. Pass@K-S serves as a more stable proxy to exploration than Pass@K. 
    Pass@K is essentially Pass@K-1 following such a definition.
    Besides Pass@K and Pass@K-S, we also track diversity of the generations using Distinct Equations proposed by~\citet{wu2024progress}, which measures the proportion of unique equations among all correct generated responses.
    
    \item \emph{Exploitation:}
    % Best-of-K accuracy measures whether the top one response ranked by the reward function is correct or not, which directly reflects how well the reward can potentially select one good response. 
    Best-of-K accuracy measures whether the top one response ranked by the reward function is correct, directly reflecting how well the reward can potentially select a good response.
    Since it is typically required to select multiple responses rather than one in self-improving training, we are interested in the reward's exploitation to select multiple responses as well. To this end, we come up with the Reward@K-S metric, which measures whether the top S candidates ranked by the reward are all correct or not. Best-of-K accuracy is a special case of Reward@K-S when S is equal to 1.
    One might think that Reward@K-S would be equivalent to Pass@K-S if only the final answer is used to select responses, potentially making Reward@K-S less useful in such cases. 
    However, we emphasize that the exploitation metrics are mainly used to measure the effectiveness when additional reward models are integrated, as we will show next in \textsection\ref{sec:dynamics} that our reward function combines final answer supervision and a reward model.
    % and Distinct Equations, reflect the model's ability in either exploitation or exploration, but fail to capture a balance between the two. Drawing inspiration from these metrics, we propose two new ones: Pass@K-S and Reward@K-S. These metrics comprehensively track the dynamic interplay between exploitation and exploration during model training, shedding light on the underlying causes of stagnation in self-training.  
\end{itemize}

Next, we conduct a case study to dive into self-improving training through tracking these metrics. 

\begin{figure*}[!t]
    \centering
    % \subfigure[Pass@1 on GSM8K]{
    %     \includegraphics[width=0.23\textwidth]{fig/gsm8k_pass@1_math_only_v2.pdf}
    % }
    \subfigure[Pass@1 on MATH]{
        \includegraphics[width=0.23\textwidth]{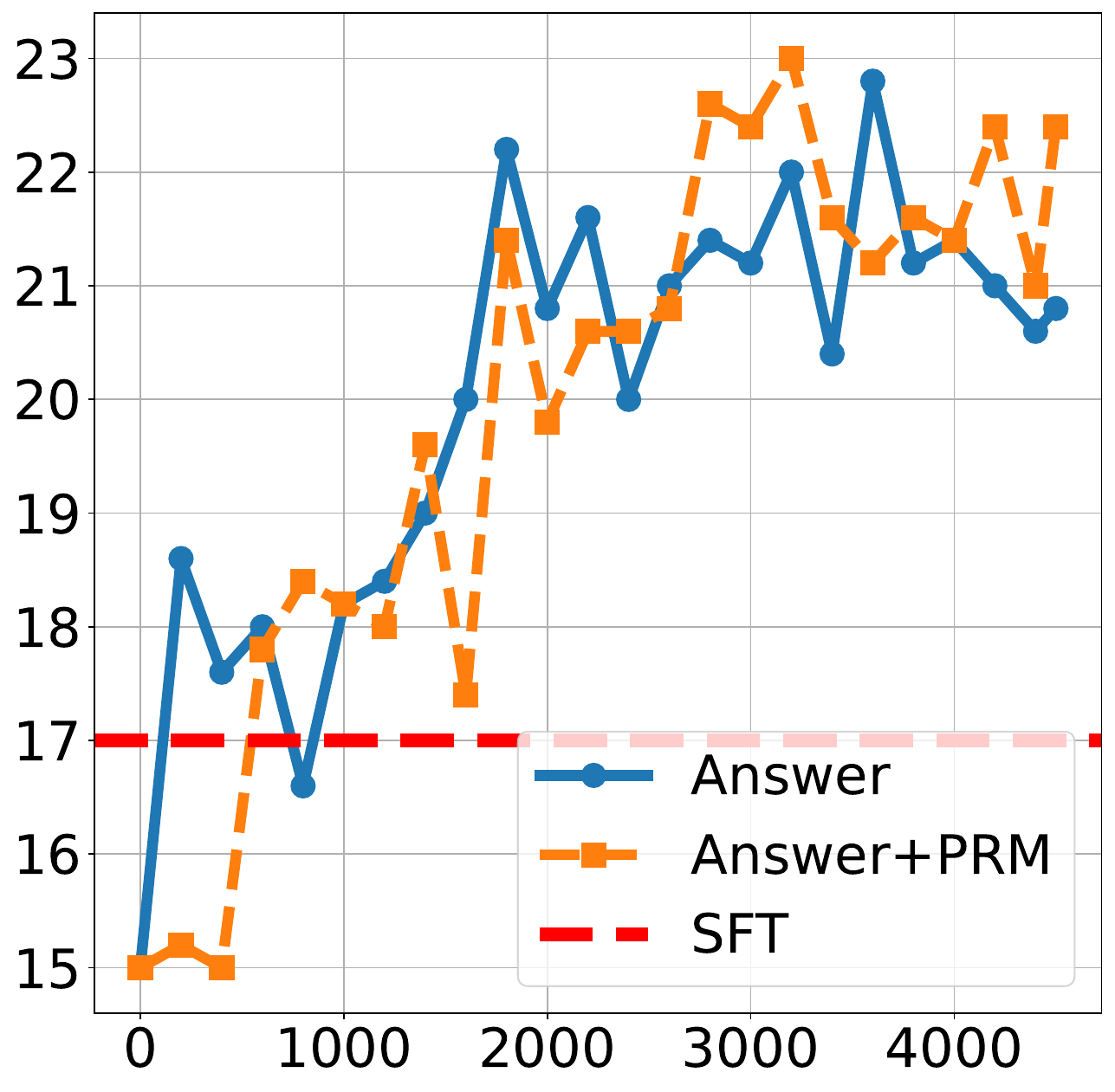}
        \label{fig:pass1_math}
    }
    % \subfigure[Diversity on GSM8K]{
    %     \includegraphics[width=0.23\textwidth]{fig/gsm8k_diversity_math_only_v1.pdf}
    % }
    \subfigure[Diversity on MATH]{
        \includegraphics[width=0.23\textwidth]{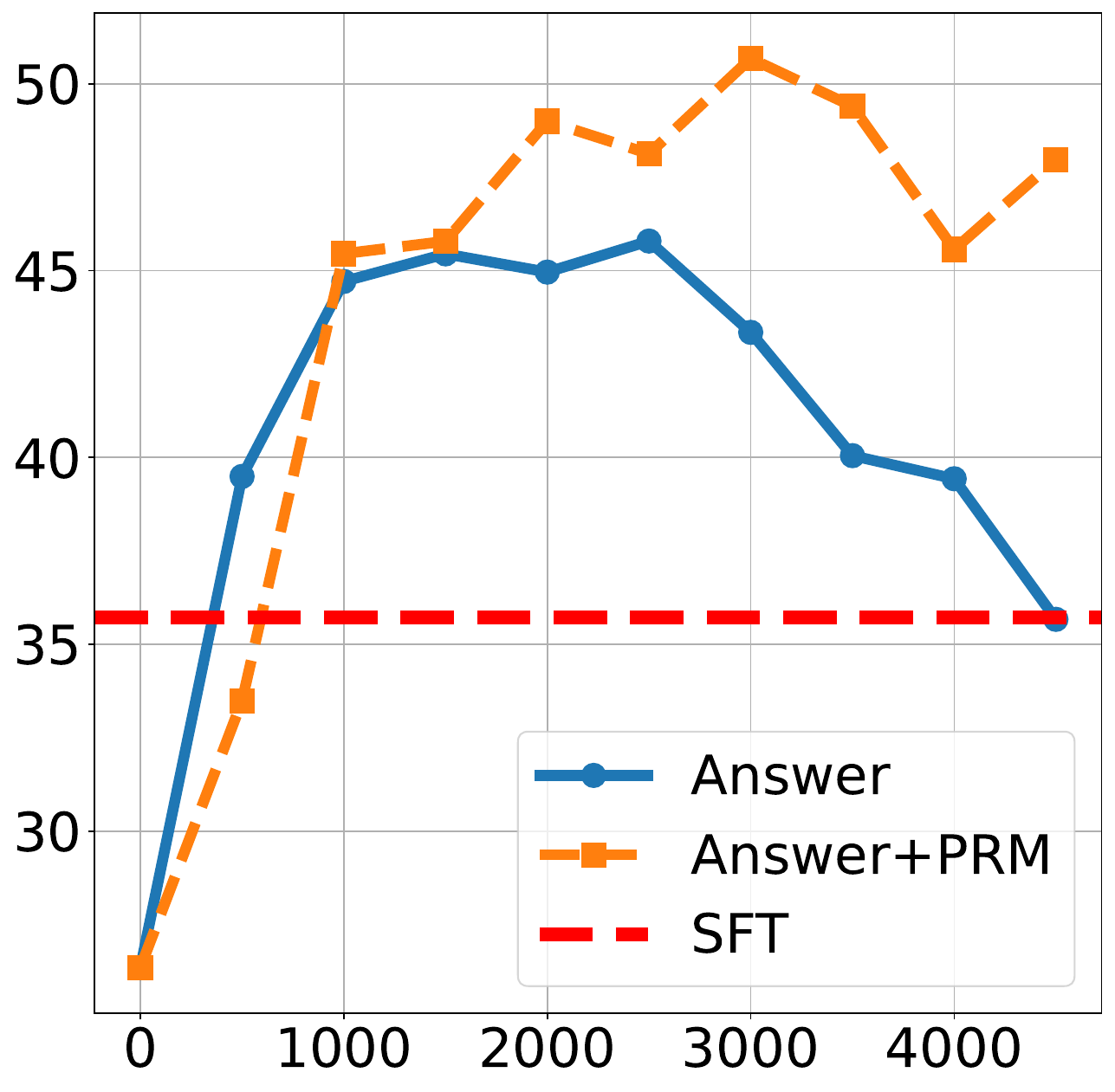}
    \label{fig:deiver_math}
    }
    %\label{fig:deiver_math}
    \subfigure[Pass@K-S on MATH]{
        \includegraphics[width=0.23\textwidth]{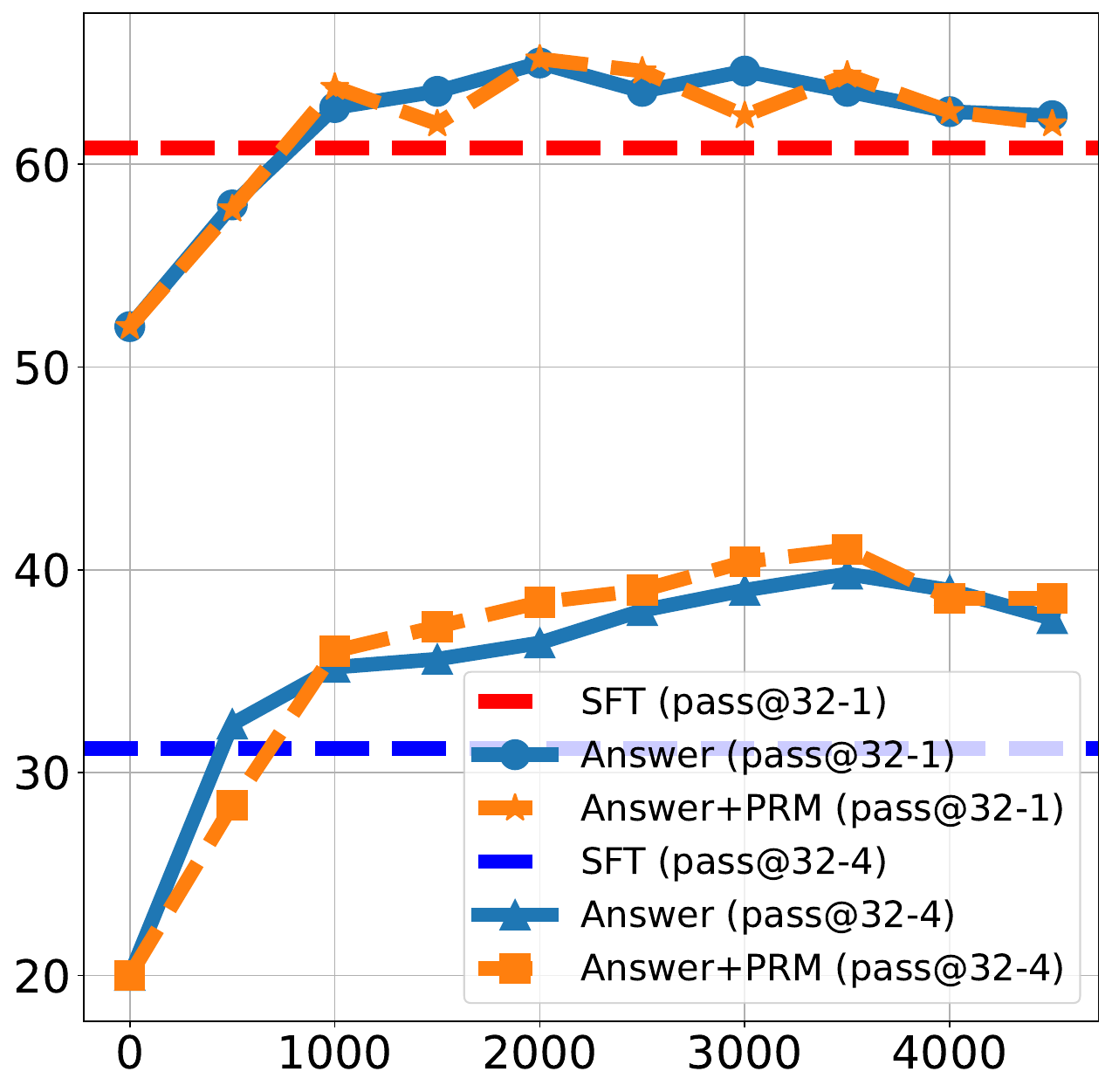}
        \label{fig:passk-s_math}
    }  
    \subfigure[Reward@K-S MATH]{
        \includegraphics[width=0.23\textwidth]{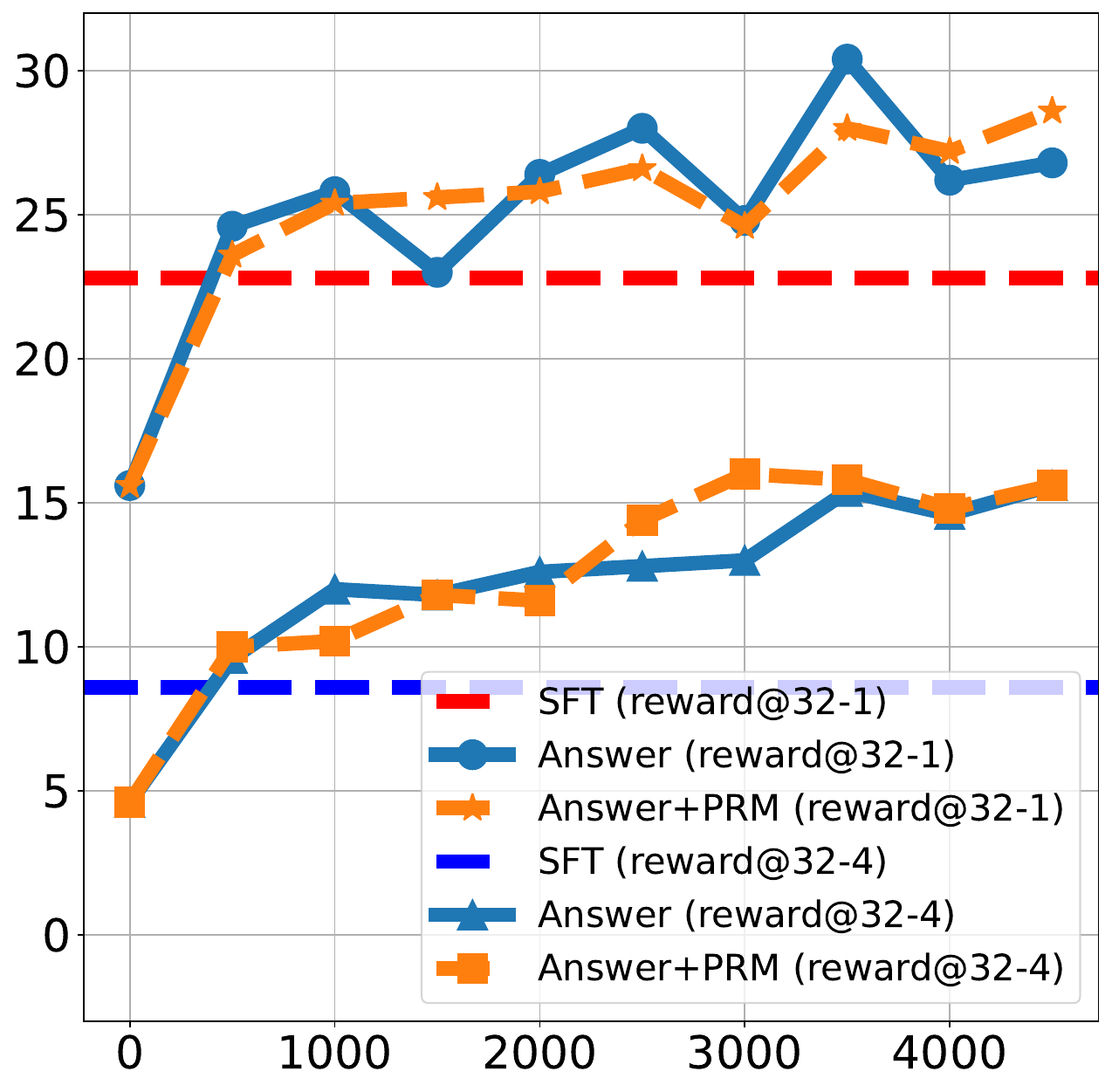}
         \label{fig:rewardk-s_math}
    }
    % \hspace{-5pt}
    \caption{Pass@1, Diversity, Pass@K-S and Reward@K-S over training steps on MATH. SFT refers to direct fine-tuning using the original dataset, "Answer" indicates matching against the ground-truth final answer, and "Answer + PRM" combines ground-truth final answer matching with PRM reward.}
    \label{fig:pass@1&diversity}
    \vspace{-10pt}
\end{figure*}

% \noindent\textbf{Self-Improvement Approaches}

\subsection{Dynamics of Exploration and Exploitation -- A Case Study in Mathematical Problem Solving}
\label{sec:dynamics}

% \jh{``Evaluation'' is expected to be moved before 2.3, and many of the other setup details below can be moved to Appendix}

% \wz{have moved the experiment setup like dataset, baseline and deitals to Appendix}

\label{sec:case_study}
In this section, we perform a case study to analyze the dynamics of exploration and exploitation in a mathematical reasoning task. Specifically, we follow~\citet{singh2023beyond} to adopt MATH ~\citep{hendrycks2021measuring} training set as the training data, evaluating on the test split of MATH. We also conduct evaluation on the test split of GSM8K, with experimental results shown in the Appendix \ref{sec:results_on_gsm8k}.

\paragraph{Setup.}
As introduced in \textsection\ref{sec:background}, we adopt the online RFT training framework in our implementation. We use Mistral-7B~\citep{jiang2023mistral} as our base model. We run the SFT baseline on MATH for 3 epochs, and use the checkpoint at the first epoch as the initial checkpoint to run self-improving training. We experiment with two different types of reward functions: 
\begin{itemize}[leftmargin=*]
    \item \emph{Answer:} we just match the predicted answer with the ground-truth final answer and keep the responses with correct answers, following~\citep{singh2023beyond}:
    \begin{equation}
    \label{eq:match_reward}
        r = \mathds{1}(\hat{a}=a^*),
    \end{equation}    
    \item \emph{Answer + PRM:} 
    %\yh{Plz match with the name in the figure} 
    we train a process reward model (PRM) using the approach in ~\citet{wang2024math} based on Mistral-7B. Then we combine the final answer reward and the PRM reward as: 
    \begin{equation}
    \label{eq:reward}
        r = \mathds{1}(\hat{a}=a^*) + r_{prm}(x, \hat{y}),
    \end{equation}
    where $\hat{a}, a^*$ are the predicted answer and the ground-truth answer respectively, $\mathds{1}(\cdot)$ is the indicator function, $\hat{y}$ is the predicted solution. $r_{prm}(\cdot)$ is the PRM score. Since PRM is designed to score every step of the solution, we choose the minimal score across all steps as the score for the entire solution, following~\citet{lightman2023let}. 
    We only keep responses with $r > \tau$ to train the model, and $\tau$ is a threshold. We found $\tau=0$ is a good hyperparameter in our early trials with different thresholds, thus we keep $\tau=0$ in all the experiments on dynamics analysis.
\end{itemize}
 
We train all methods for 4500 training steps. For our online RFT, we adopt an iteration interval of 500 steps, resulting in a total of 9 iterations -- significantly higher than the typical 3–5 iterations reported in prior work~\citep{singh2023beyond}. We sample 32 candidate solutions per query during training with a temperature of 1.0. 
We include the SFT baselines, which are trained for 3 epochs on MATH. Please see Appendix~\ref{sec:casestudy_setup} for more setup details.

\paragraph{Observation 1: Exploration decreases over training, and PRM helps retain the exploration ability.}
% \textbf{Decline in Diversity}
As demonstrated in Figure~\ref{fig:pass1_math},
the self-improvement method significantly outperforms SFT in enhancing the model's ability to generate accurate responses through greedy decoding. However, for both methods, the performance improvements diminish over time as the number of iterations increases. From Figure~\ref{fig:deiver_math}, the diversity metric decreases dramatically, a similar phenomenon is observed in~\citet{wu2024progress}. 
Notably, the Answer+PRM reward is able to overcome the declining trend and retain exploration. 
We hypothesize that filtering responses solely based on answer correctness often results in homogeneous reasoning paths, whereas the fine-grained reward strategy encourages the selection of high-quality paths, thereby preserving diversity.
Examining the Pass@K-S metrics in Figure~\ref{fig:passk-s_math}, Pass@K-S initially increases, indicating an improvement in model exploration. However, Pass@K-S subsequently declines, with Pass@K-1 accuracy falling close to the SFT baseline.
%While integrating PRM slightly improves Pass@K-S on GSM8K, its impact on MATH is limited, also due to the limited capability of the reward model. 
The plateau pattern of exploration is not a good sign, as it suggests that the model may not learn new things to explore better. Since the reward model remains fixed during training, the model's ability stagnates once exploration halts.
% The Self-Improvement method tends to cause a rapid decline in solution diversity. However, we found that using a more fine-grained Combination Reward can effectively mitigate this issue. As shown in Figure~\ref{fig:pass@1&diversity}, the diversity decreases more gradually with the Combination Reward compared to the Sparse Reward. This is because filtering solutions solely based on answer correctness often leads to many homogeneous results. In contrast, the fine-grained reward strategy promotes the selection of high-quality steps, which indirectly helps preserve solution diversity.

% \textbf{Observation 3: Exploitation saturates on GSM8K while keeps improve on MATH:}
\textbf{Observation 2: Exploitation keeps improving on MATH.} 
As shown in Figure~\ref{fig:rewardk-s_math}, we observe Reward@K-S continues to improve on MATH, potentially because we are training with the MATH dataset. 
Since our our reward remains fixed during training, the exploitation performance is closely related to exploration of the policy model, which is a moving target as discussed in \textsection\ref{Sec:factor}.
% Notably, our reward remains static during training, as we do not update it or alter how we select responses. Therefore, exploitation performance is closely related to exploration by the policy model, which is a moving target. 
However, as shown in our previous analysis, the model’s exploration does not continually improve in all cases. We hypothesize that this is the key factor that bottlenecks self-improving training. 
Exploration is related to configurations such as how we sample responses from the model, how many samples to draw. Similarly, exploitation depends on how we utilize the reward to select responses. While all previous works have treated these configurations as static during training, as far as we know, the question arises: Can we adjust them dynamically to better align exploration and exploitation? We investigate this question next.
% While all previous works treat these configurations as static during training as far as we know, can we adjust them dynamically so that exploration and exploitation better fit each other? We study this question next.

% As iterations progress, the model's ability to generate diverse and accurate solutions stagnates or even significantly declines. For example, in Figure~\ref{fig:pass@k-s&reward@k-s}, the pass@32-1 and pass@32-4 metrics for Sparse Reward sharply decrease after 4000 steps, eventually falling below those of the SFT model. This decline hampers subsequent iterations because the model begins to produce more homogeneous solutions, with little improvement in quality. As a result, the post-training data for the next iteration suffers, further affecting the model's performance.

% \textbf{Reward Model Bias Emerges} As the model continues to optimize, it becomes increasingly challenging for the reward model to select accurate solutions. For instance, in Figure~\ref{fig:pass@k-s&reward@k-s}, after 6,000 steps, the improvements in reward@32-1 and reward@32-4 slow down and even begin to decline. This suggests that the existing reward model develops a bias in estimating rewards for the model's solutions, making it less effective in supervising further progress~\citep{shao2024deepseekmath}. Therefore, timely updates to the reward model are essential to ensure continuous improvement in the training process.

% TODO
% 1. Pass@1 for GSM8K and MATH
% 2. Distinct Equations for GSM8K and MATH
% 3. Pass@K-S for GSM8K and MATH
% 4. Reward@K-S for GSM8K and MATH

\section{\bs{} -- Balanced Self-Taught Reasoners}
\label{sec:unified_view}

% In this section, we introduce the Query Effect metric, which measures the positive impact of post-training data during the self-improvement process. We then analyze how Query Effect changes dynamically in existing self-improvement methods. Finally, we explore the factors that may influence Query Effect.

In this section, we firstly introduce a metric to evaluate the interplay effect between exploration and exploitation. Next, we analyze its relationship with different configurations. Finally, we present our complete algorithm, which automatically adjusts the exploration and exploitation abilities. 
% the query effect metric, designed to assess a query's potential in relation to the current model's exploration and exploitation capabilities. We start by examining how the query effect fluctuates within existing self-improvement methods. Finally, we explore the factors associated with exploration and exploitation that may impact the query effect.

% \begin{figure*}[t]
%     \centering
%     % \subfigure[Sample Size]{
%     %     \includegraphics[width=0.30\textwidth]{fig/qe&step.pdf}
%     %     \label{fig:sample}
%     % }
%     \subfigure[Best Temperature]{
%         \includegraphics[width=0.30\textwidth]{fig/temp_iter_bar.pdf}
%         \label{fig:temp}
%     }
%     \subfigure[Best Threshold]{
%         \includegraphics[width=0.30\textwidth]{fig/thre_iter_bar.pdf}
%         \label{fig:threshold}
%     }
%     \subfigure[Sample Size]{
%         \includegraphics[width=0.30\textwidth]{fig/qe_step.pdf}
%         \label{fig:sample}
%     }
%     % \vspace{-5pt}
%     \caption{{\bf Left:} the best sampling temperature to achieve the max average balance scores at different training steps; {\bf Middle:} the best reward threshold to achieve the max average balance scores at different training steps; {\bf Right:} the average balance scores on 600 MATH training queries at different training steps, varying the number of samples to draw per query; Every 500 steps corresponds to one iteration - for example, step 500 corresponds to the 1 st iteration, while step 1500 corresponds to the 3 rd iteration.
%     }
%     \label{fig:qe_and_hyperparam}
%     % \vspace{-10pt}
% \end{figure*}

\subsection{Balance Score}
\label{sec:query-effect}
% Section~\ref{sec:ana_self_improve} highlights that the model’s capacity to consistently generate accurate and diverse candidate solutions, along with the reward model’s ability to reliably select high-quality ones, are key to achieving self-improvement. From the perspective of post-training data, both of these abilities serve the same purpose: ensuring that the positive impact of the data used for improvement in each iteration is continuously strengthened.

% \jh{I have been feeling ``query effect'' is not a good name since long time ago because it sounds like a metric to evaluate quality of queries, but our metric is more like quality of the responses. I am thinking sth like ``response quality'' or ``synthesis quality'', what do you think?}\yh{Or response utility. I think we should avoid ``quality'' in its name, which make the discussion about quality and quantity confusing.  }
\textsection\ref{sec:ana_self_improve} highlights the importance of a model's exploration and exploitation capabilities for self-improvement, emphasizing their continuous evolution during training. Now we seek a metric that could capture the interplay of these two factors.
Intuitively, we hope the model can explore a diverse range of high-quality responses and with the reward function distinguishing between them.
Specifically, given a query, two conditions to be satisfied for the selected responses to effectively contribute to training: (1) a substantial absolute number of high-quality responses, and (2) a high proportion of high-quality responses among the selected ones. The first condition ensures there is a sufficient quantity of high-quality data for training, while the second prioritizes maintaining a high proportion of good responses to prevent contamination of the training set with low-quality data. 
For example, selecting only two correct responses may result in a perfect 100\% quality ratio but provide insufficient data for training. Conversely, selecting all 64 candidates -- of which 16 are correct -- yields a higher absolute number of high-quality responses but lowers the quality ratio to 25\%, introducing too many incorrect responses into the training process. Therefore, these two capture different aspects of the data.
% (1) The number of high-quality responses among the selected ones cannot be too small, otherwise we do not have enough good data for training; (2) The percentage of the high-quality responses among the selected ones must be large enough, otherwise we may mix too many poor-quality data into the training data. 
% The former focuses on the quantity of high-quality responses selected, whereas the latter prioritizes the ratio (but not the absolute quantity) of the high-quality responses among all selected responses. 
% For example,  if we select only 2 responses and they are correct, then the ratio is 100\% while the absolute number is only 2, which cannot give us enough training data; if we select all 64 candidates, suppose 16 of them are correct and others are wrong, then we have 16 high-quality responses, but the ratio is only 25\% -- feeding too many incorrect responses into training is undesirable. Therefore, these two capture different aspects of the data.
Following these two intuitions, we propose a metric, \emph{balance score}, to measure the balance effect -- or in other words, the overall contribution of the synthetic data -- given the exploration- and exploitation-related configurations. We detail it below. 

For each query $x_{i}$, let $n_{i}$ represent the total number of selected responses for it, and $n_{i}^{\prime}$ denote the number of unique, correct responses among them, then the balance score is defined as follows:
\begin{equation} \label{eq:query_effect}
bs_i = \min \left( \frac{n_{i}^{\prime}}{n^{\star}},\ 1 \right) \cdot \frac{n_{i}^{\prime}}{n_{i}},
\end{equation}
where $(\frac{n_{i}^{\prime}}{n^{\star}},\ 1)$ is a discount factor that encourages the number of correct solutions to be larger than a pre-specified parameter $n^{\star}$ -- there is no discount when we have more than $n^{\star}$ correct responses. We impose a cap of 1, rather than encouraging $n_{i}^{\prime}$ to be as large as possible, because otherwise the number of responses among queries will be severely imbalanced~\citep{tong2024dart}, where the easy queries will occupy most of the correct responses to maximize the average balance score. The second term, $n_{i}^{\prime} / n_{i}$, is the ratio of the correct responses. 
We note that this ratio is always 100\% if only the final-answer reward is used. $n^{\star}$ is the only hyperparameter in this metric, and it roughly implies the number of correct responses that we aim to have per query.
Suppose we aim to select $N$ samples per iteration, and each iteration we feed in $M$ queries where $N > M$, then we simply decide $n^{\star} = \left\lceil \frac{N}{M} \right\rceil$. As $N$ and $M$ are just general training hyperparameters related to data loader, we never tune them in the paper and just set them as reasonable numbers as we will describe later. This means, \emph{the balance score does not introduce additional hyperparameters for us to tune}. 
% \jh{In setup section 4.1, we should explicitly mention our M and N in the main experiments}

Intuitively, the exploration and exploitation are desired to be controlled to maximize the average balance score -- the max value of $bs_i$ is 1 and the average balance score is maximized when all the selected responses are correct and there are at least $n^{\star}$ correct responses for each query. Controlling exploration and exploitation is non-trivial and there could be different approaches, in this work we focus on a simple method -- where we manipulate the balance through certain hyperparameter configurations such as the sampling temperature and reward threshold, which we discuss next.

\begin{figure*}[t]
    \centering
    % \subfigure[Sample Size]{
    %     \includegraphics[width=0.30\textwidth]{fig/qe&step.pdf}
    %     \label{fig:sample}
    % }
    \subfigure[Best Temperature]{
        \includegraphics[width=0.30\textwidth]{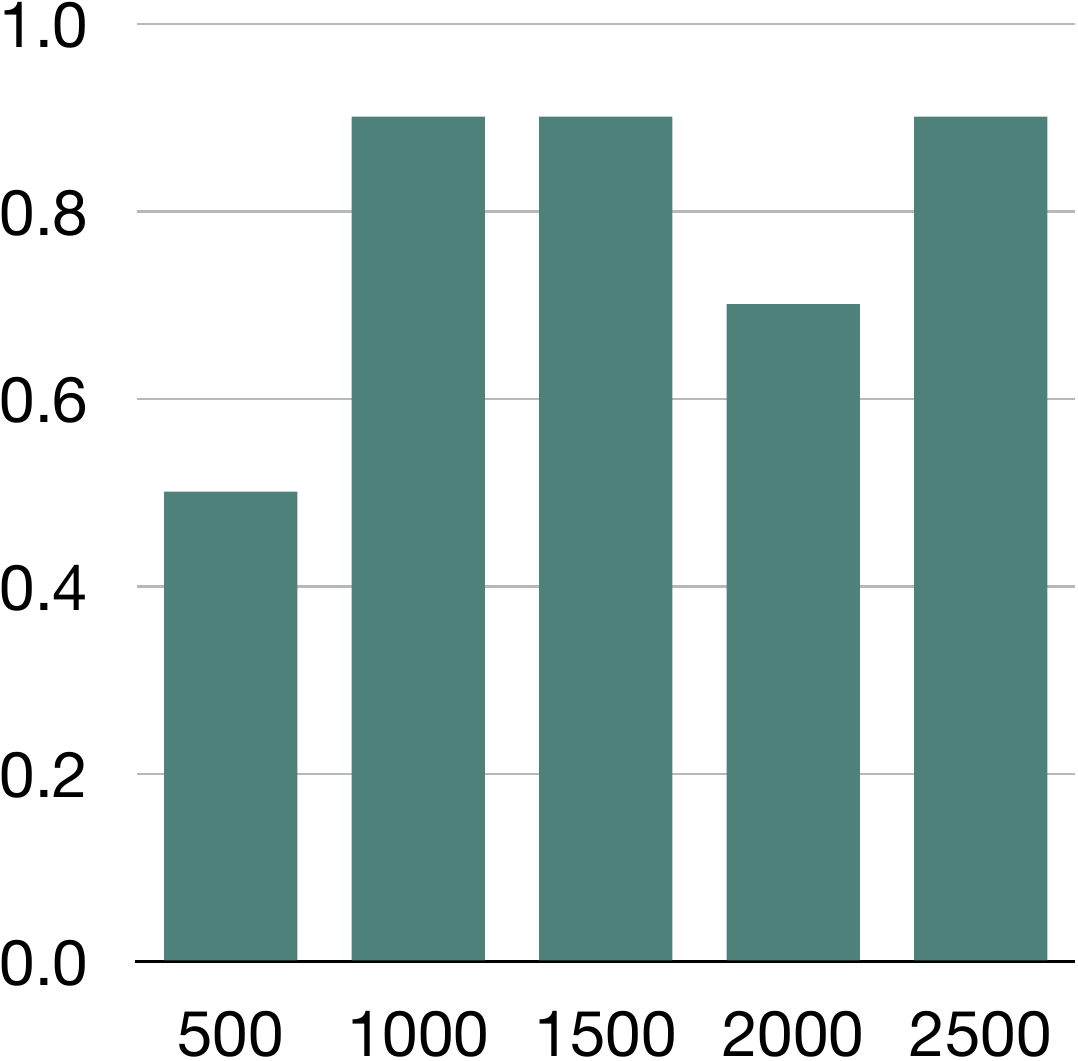}
        \label{fig:temp}
    }
    \subfigure[Best Threshold]{
        \includegraphics[width=0.30\textwidth]{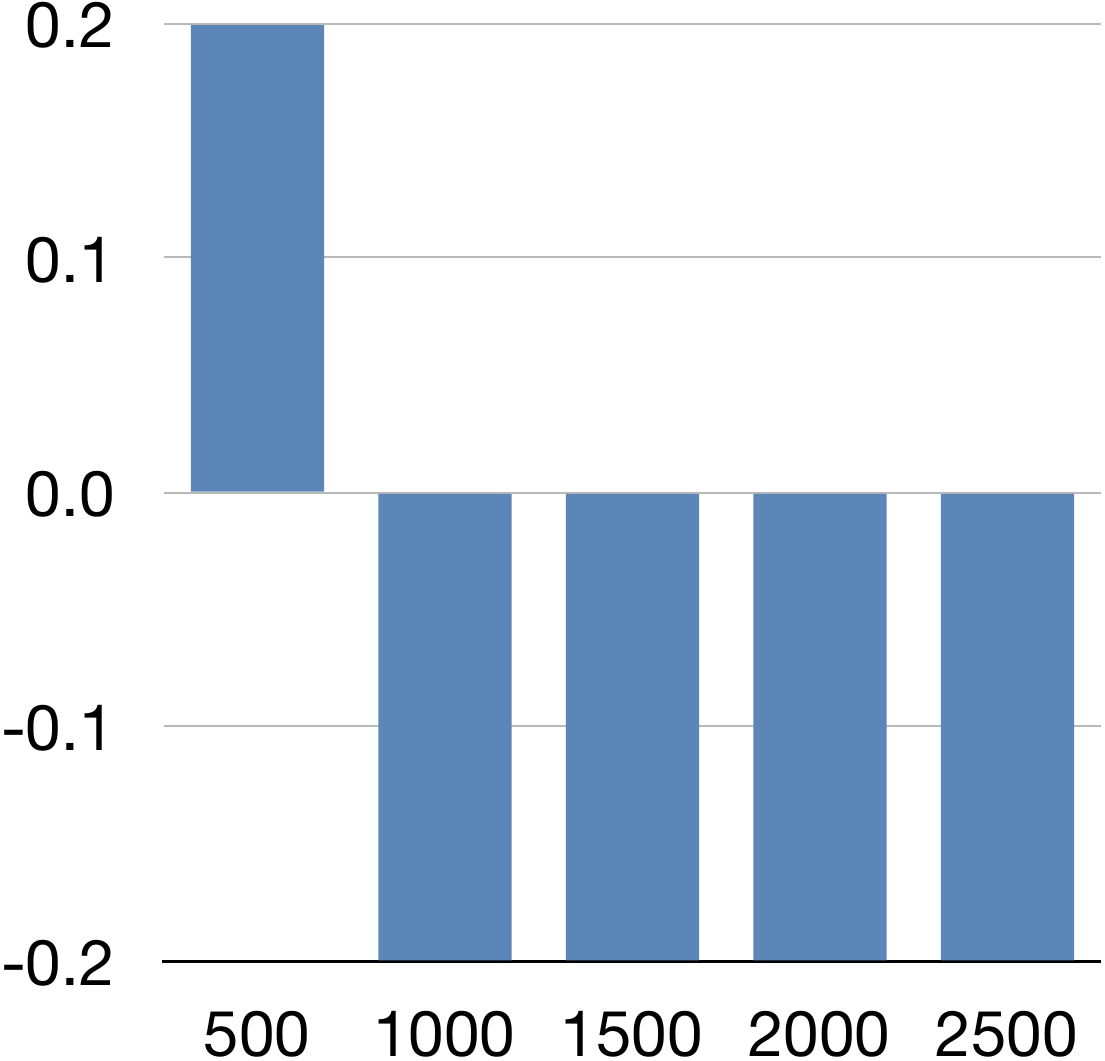}
        \label{fig:threshold}
    }
    \subfigure[Sample Size]{
        \includegraphics[width=0.30\textwidth]{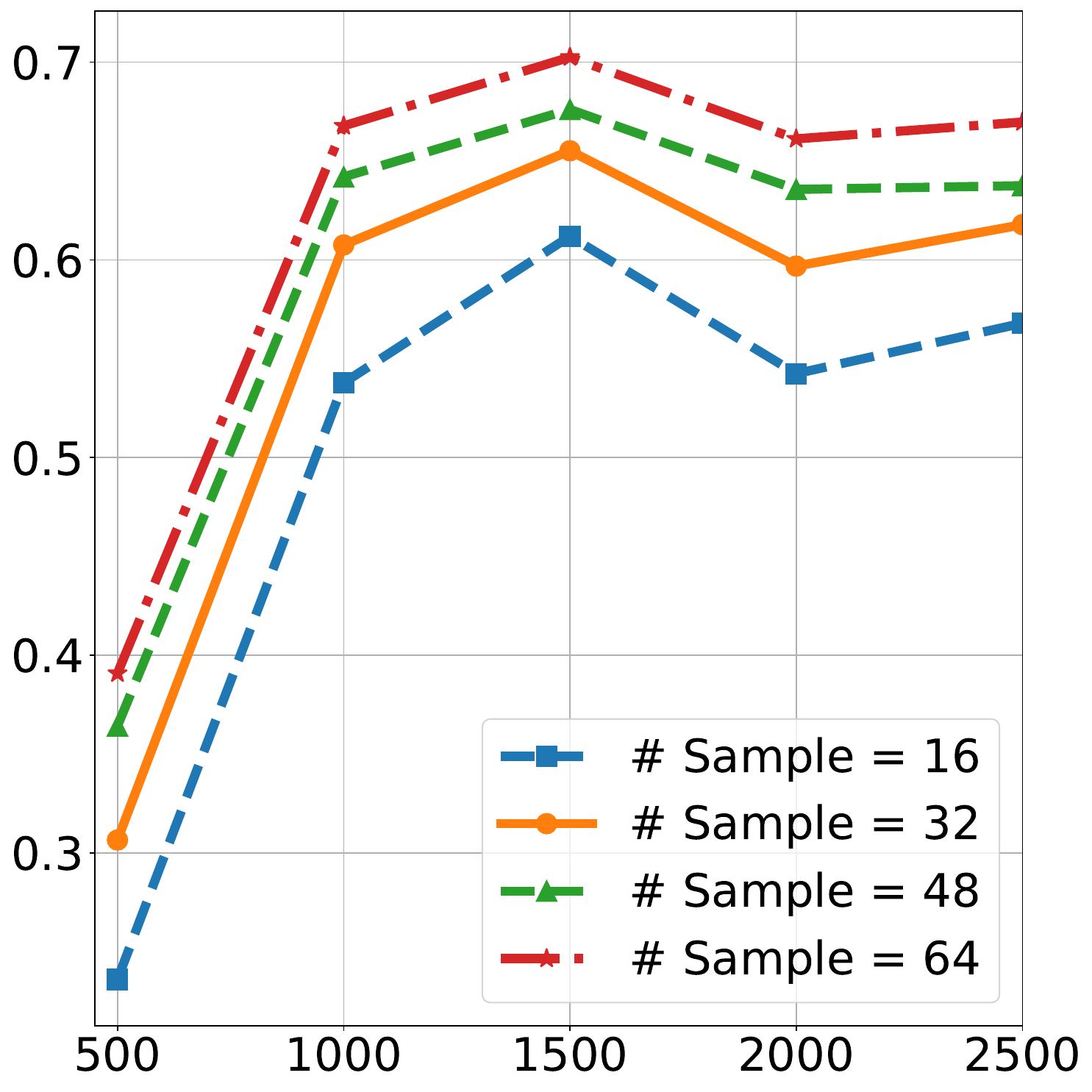}
        \label{fig:sample}
    }
    % \vspace{-5pt}
    \caption{{\bf Left:} the best sampling temperature to achieve the max average balance scores at different training steps; {\bf Middle:} the best reward threshold to achieve the max average balance scores at different training steps; {\bf Right:} the average balance scores on 600 MATH training queries at different training steps, varying the number of samples to draw per query; Every 500 steps corresponds to one iteration - for example, step 500 corresponds to the 1 st iteration, while step 1500 corresponds to the 3 rd iteration.
    }
    \label{fig:qe_and_hyperparam}
    % \vspace{-10pt}
\end{figure*}

\subsection{Dynamic or Static -- On the Configurations of Exploration and Exploitation}
\label{sec:config-analysis}

% \begin{figure}[!t]
%         \centering
%         \includegraphics[width=0.95\columnwidth]{fig/b_star_long.pdf}
%         \caption{Illustration of the \bs{} approach. In each iteration, \textbf{Balance Exploration and Exploitation}: we first identify the configurations—temperature from set $\mathcal{T}$ and reward threshold from set $\Theta$—that maximize the average query effect using a small subset of training queries. Next, \textbf{Generating and Rewarding}: we apply the optimal temperature and threshold to generate and reward the full set of training queries. Finally, \textbf{Improving}: we update the model based on the selected data.\jh{this should appear as figure 2}
%         }
%         \label{fig:main}
%     \vspace{-10pt}
% \end{figure}

\label{sec:config_query_effect}

The balance score provides a straightforward signal on how we should manipulate exploration and exploitation to be optimal. In this work, we mainly focus on adjusting some hyperparameter configurations to control exploration and exploitation. Here, we conduct a preliminary analysis to investigate whether the optimal configurations -- which maximize the averaged balance score -- are static or should be dynamically changing.
% We explore various configurations that directly relate to exploration and exploitation. 
% , focusing on both exploration-related settings, such as sampling temperature and sample size, and exploitation-related settings, like reward thresholds. 
% Specifically, we obtain the policy model checkpoints and reward model checkpoints of different iterations from the online RFT with final Answer + PRM reward run, 
Specifically, we obtain the policy model checkpoints and reward model checkpoints from different iterations of online RFT with the final Answer + PRM reward run, with 500 steps between each iteration. We then apply different configurations to these checkpoints, and compute the average balance scores on 600 randomly sampled MATH training queries. Below we detail the configurations we study.

\paragraph{Exploration -- temperature.}
% \jh{if we remove sample size from the first paragraph, some sentences here may need to be revised a bit}We next examine how sampling temperature affects query effect. 
Sampling temperature, an important configuration influencing exploration, is first examined for its impact on balance score.
With the reward threshold fixed at 0.0 and a sample size of 32, we adjust the sampling temperature to 0.5, 0.7, 0.9, and 1.1. In Figure~\ref{fig:temp} we show the optimal temperature we obtained that maximizes the average balance score at different iterations (training steps). In this particular setting, lower temperatures are preferred in the beginning while higher temperatures are better later on.  
% that while sampling temperature does not have a consistent, monotonic effect on query effect, we still observe notable trends. \textbf{Early in training, lower temperatures yield better query effect, whereas as training progresses, higher temperatures become necessary to maintain or improve performance.} 
This phenomenon can be explained by the model’s shifting limitations during training. In the early stages, the model's ability to generate correct solutions is limited, so lower temperatures help sample more accurately~\citep{yang2023large}. As training advances, the challenge shifts to preserving diversity in the generated solutions, requiring higher temperatures to ensure broader sampling.

\noindent\textbf{Exploitation -- threshold.} 
We investigate the impact of reward thresholds on balance score, which determines how the reward exploits the responses. 
A high reward threshold indicates that the reward model strictly exploits responses, whereas a low threshold suggests more relaxed selection criteria. In our experiments, we fix the sampling temperature at 1.0 and the sample size at 32, while varying the reward threshold and selecting responses that exceed the threshold. Figure~\ref{fig:threshold} presents the optimal threshold that maximizes the average balance score.
% , show that \textbf{reward thresholds influence the query effect differently across datasets}. 
It indicates that a higher threshold is preferred in the beginning, but it may need to decrease as training progresses. Intuitively, this suggests that more rigorous filtering should be applied in the beginning, when the model is weaker, and the threshold should be relaxed slightly as the model becomes stronger.

\subsection{\bs{}}

% \begin{figure}[!t]
%         \centering
%         \includegraphics[width=0.95\columnwidth]{fig/b_star.pdf}
%         \caption{Illustration of the \bs{} approach. In each iteration, we first identify the configurations—temperature from set $\mathcal{T}$ and reward threshold from set $\Theta$—that maximize the average query effect using a small subset of training queries (top left). Next, we apply the optimal temperature and threshold to generate and reward the full set of training queries (bottom). Finally, we update the model based on the selected data (top right).
%         }
%         \label{fig:main}
%     \vspace{-10pt}
% \end{figure}

Building on the findings from \textsection \ref{sec:config-analysis} that the optimal configurations to maximize balance score is dynamically changing, we propose \bs{}, short for Balanced Self-Taught Reasoners, a method that maximizes the average balance score by dynamically adjusting configurations to balance exploration and exploitation. We also include related work on dynamically adjusting configurations in the Appendix \ref{sec:related_work}.
Specifically, we adjust temperature and threshold automatically at every iteration, to maximize the average balance score. 
%To demonstrate that \bs{}'s improvements stem from dynamic configuration adjustments rather than from suboptimal configuration settings, we compare \bs{} with Online RFT using various fixed configuration combinations in the Appendix \ref{sec:fix_config}.
Notably, we only need to compute the balance score on a small subset of training queries to decide the balanced configurations, thus incurring negligible additional costs compared to the baselines. For example, we only use 600 MATH queries in our experiments. Our \bs{} algorithm is illustrated in Figure~\ref{fig:main} and summarized in Algorithm~\ref{alg_bstar}. There are other configurations affecting balance score, such as the number of responses drawn per query (sample size) that will influence the exploration. In our preliminary trials in Figure~\ref{fig:sample}, larger sample size tends to be generally helpful. Therefore, we just use the maximum sample size according to our computing budget and fix it, and mainly focus on dynamic adjustments of temperature and threshold.

\begin{table*}[t]
\centering
\resizebox{0.98\textwidth}{!}{
\begin{tabular}{lcccccccccc}
\toprule
\multicolumn{1}{c}{\multirow{2}{*}{Methods}} & \multicolumn{3}{c}{GSM 8K}   & \multicolumn{3}{c}{MATH}    & \multicolumn{3}{c}{APPS}                               & \multicolumn{1}{c}{ARC-C}  \\ \cmidrule{2-11} 
\multicolumn{1}{c}{}                                        & P@1 & P@32 & P@32-4 & P@1 & P@32 & P@32-4 & P@1           & P@32          & P@32-4         & P@1 \\ \midrule
SFT                                    & 36.6   & 88.5    & 62.2      & 17.0   & 60.8    & 31.2     & 9.3              & 43.5             & 25.5             & —            \\
Rest-EM (w/o RM)                                             & 40.5   & 89.9    & 69.8      & 22.8   & 60.0    & 33.6     & 14.5             & 43.9             & 28.2             & 70.7 \\
Rest-EM (w/ RM)                                           & 46.3   & 90.7    & 72.2      & 24.2   & 62.8    & 37.4     & — & —  & —  & —            \\
Iterative RFT (w/o RM)                                       & 42.8   & 88.9    & 71.3      & 24.2   & 63.4    & 38.2     & 15.2             & 44.3             & 28.0             & 70.3 \\
Iterative RFT (w/ RM)                                     & 46.6   & 90.2    & 74.9      & 24.4   & 62.6    & 39.0     & — & —  & —  & —            \\
Online RFT (w/o RM)                                 & 44.0   & 88.1    & 69.7      & 23.0   & 57.2    & 38.2     & 17.3             & 45.8             & 27.8             & 71.2 \\
Online RFT (w/ RM)                        & 46.8   & 91.4    & 76.5      & 23.2   & 62.6    & 39.2     & —  & —  & —  & —            \\ \rowcolor{gray!25}
\bs{}                                                     & \textbf{53.8}   & \textbf{93.6 }   & \textbf{81.0}      & \textbf{27.8}   & \textbf{67.2}    & \textbf{42.2}     & \textbf{19.6}             &\textbf{49.3}             & \textbf{30.7}        & \textbf{73.0} \\ \bottomrule
\end{tabular}
}
% \vspace{-5pt}
\caption{Comparison of self-improvement methods across MATH, GSM8K, APPS and ARC-Challenge. Methods include variants with and without a reward model ("w/ RM" and "w/o RM"). The results are based on the Mistral-7B model except for APPS that is from Llama-3-8B.}
\label{tab:b_star_main_result}
\vspace{-10pt}
\end{table*}

% \begin{table*}[t]
% \centering
% \resizebox{0.8\textwidth}{!}{
% \begin{tabular}{lccccccccc}
% \toprule
% Step              & \textbf{500} & \textbf{1000} & \textbf{1500} & \textbf{2000} & \textbf{2500} & \textbf{3000} & \textbf{3500} & \textbf{4000} & \textbf{4500} \\ \midrule
% Temperature       & 0.5          & 0.8           & 0.9           & 1             & 1.1           & 1.1           & 0.9           & 1.1           & 1.1           \\
% Reward threshold & 0            & -0.1          & -0.1          & -0.1          & -0.1          & -0.1          & -0.1          & -0.1          & -0.1          \\
% Balance Score      & 0.470        & 0.538         & 0.589         & 0.621         & 0.646         & 0.660         & 0.673         & 0.678         & 0.679         \\ 
% \bottomrule
% \end{tabular}
% }
% %\vspace{-5pt}
% \caption{Dynamic configuration adjustments by \bs{} in mathematical problem-solving. The temperature increment and reward threshold increment are both set to 0.1. Additionally, finer-grained increments for these parameters are explored in detail in Appendix~\ref{sec:fine_config} and summarized in Table~\ref{tab:b_star_hyper_fine}.}
% \label{tab:b_star_hyper}
% \vspace{-10pt}
% \end{table*}

\section{Main Experiments}
\subsection{Setup}

\label{sec:bstar_setup}
We evaluate \bs's effectiveness in enhancing self-improvement for mathematical problem-solving, coding challenges and commonsense reasoning. Our evaluation compares several baseline methods: STaR/ResT-EM ~\citep{zelikman2022star,singh2023beyond}, Iterative RFT, and Online RFT~\citep{shao2024deepseekmath}.
%In contrast, Iterative RFT builds on the previous iteration by inheriting the checkpoint and continuing training from that point. Online RFT extends this further by inheriting both the checkpoint and the full training state, enabling seamless progress across iterations. 
Additionally, we implement these methods with two types of reward function:  "without RM" (Answer) and "with RM" (Answer+PRM), as described in \textsection\ref{sec:case_study}. We also provide detailed baseline methods and experimental setup in Appendix ~\ref{sec:main_setup}.

\begin{figure*}[t]
    \centering
    \subfigure[GSM8K]{
        \includegraphics[width=0.27\textwidth]{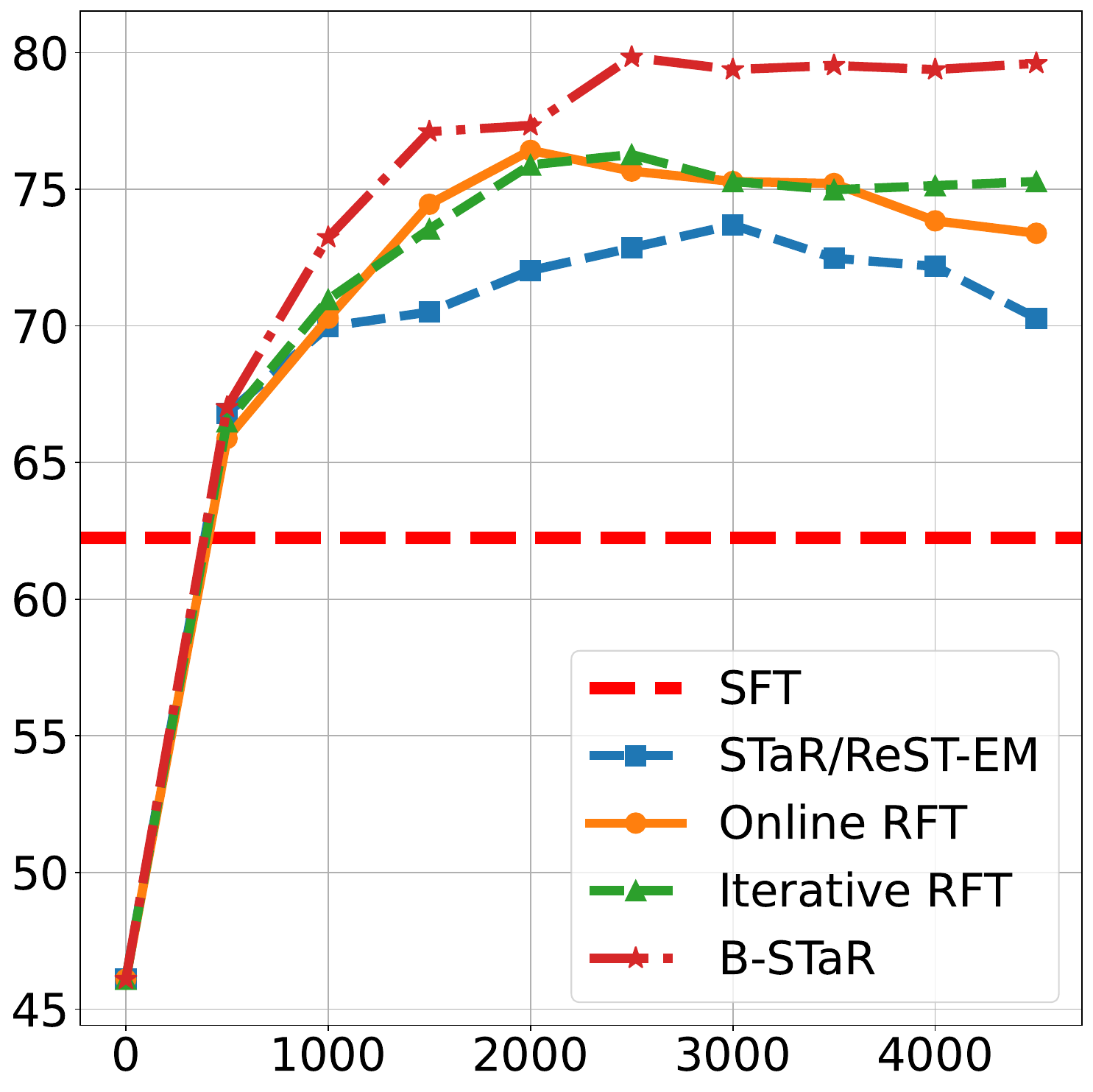}
    }
    \hspace{15pt}
    \subfigure[MATH]{
        \includegraphics[width=0.27\textwidth]{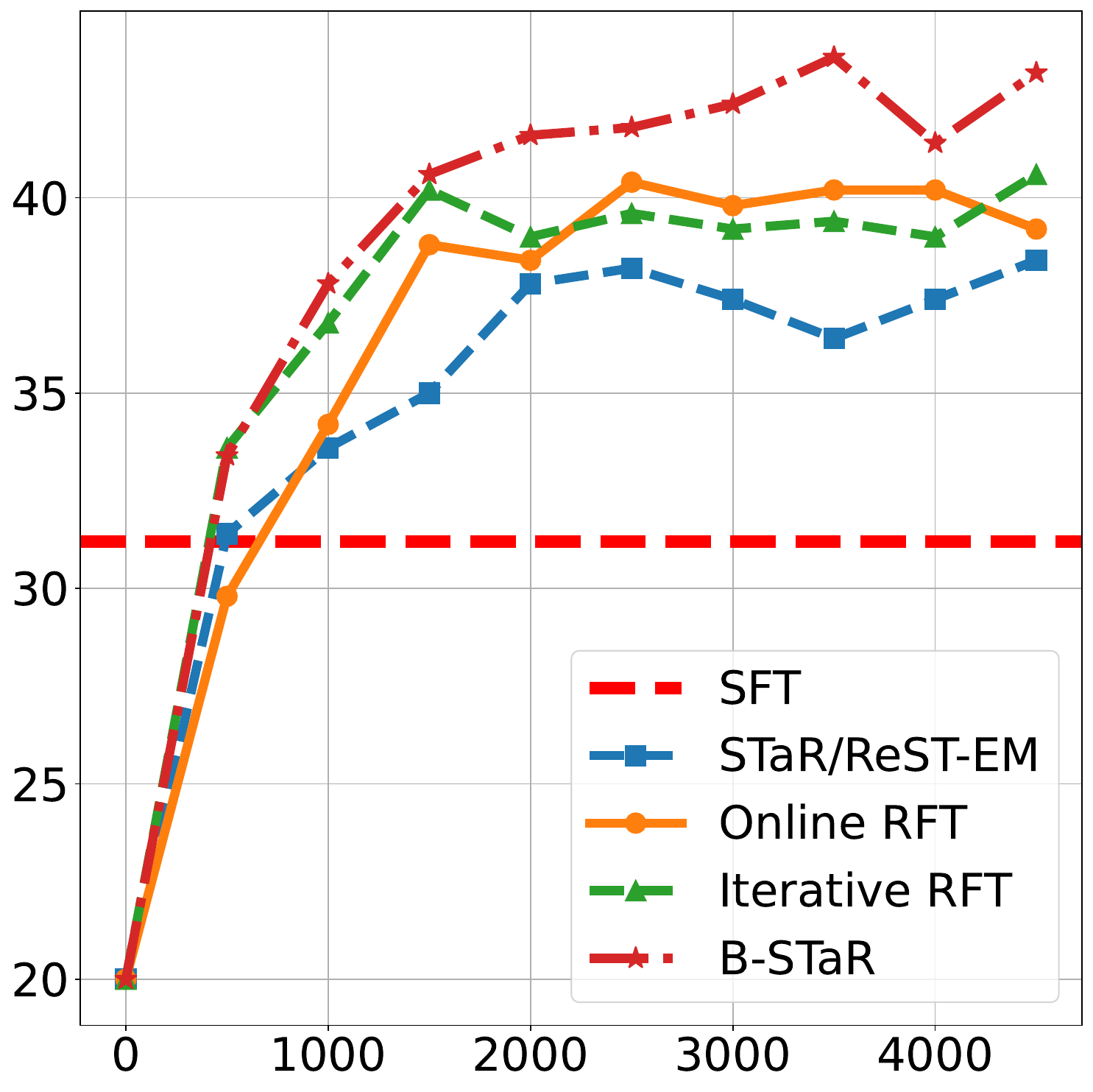}
    }
    \hspace{15pt}
    \subfigure[APPS]{
        \includegraphics[width=0.27\textwidth]{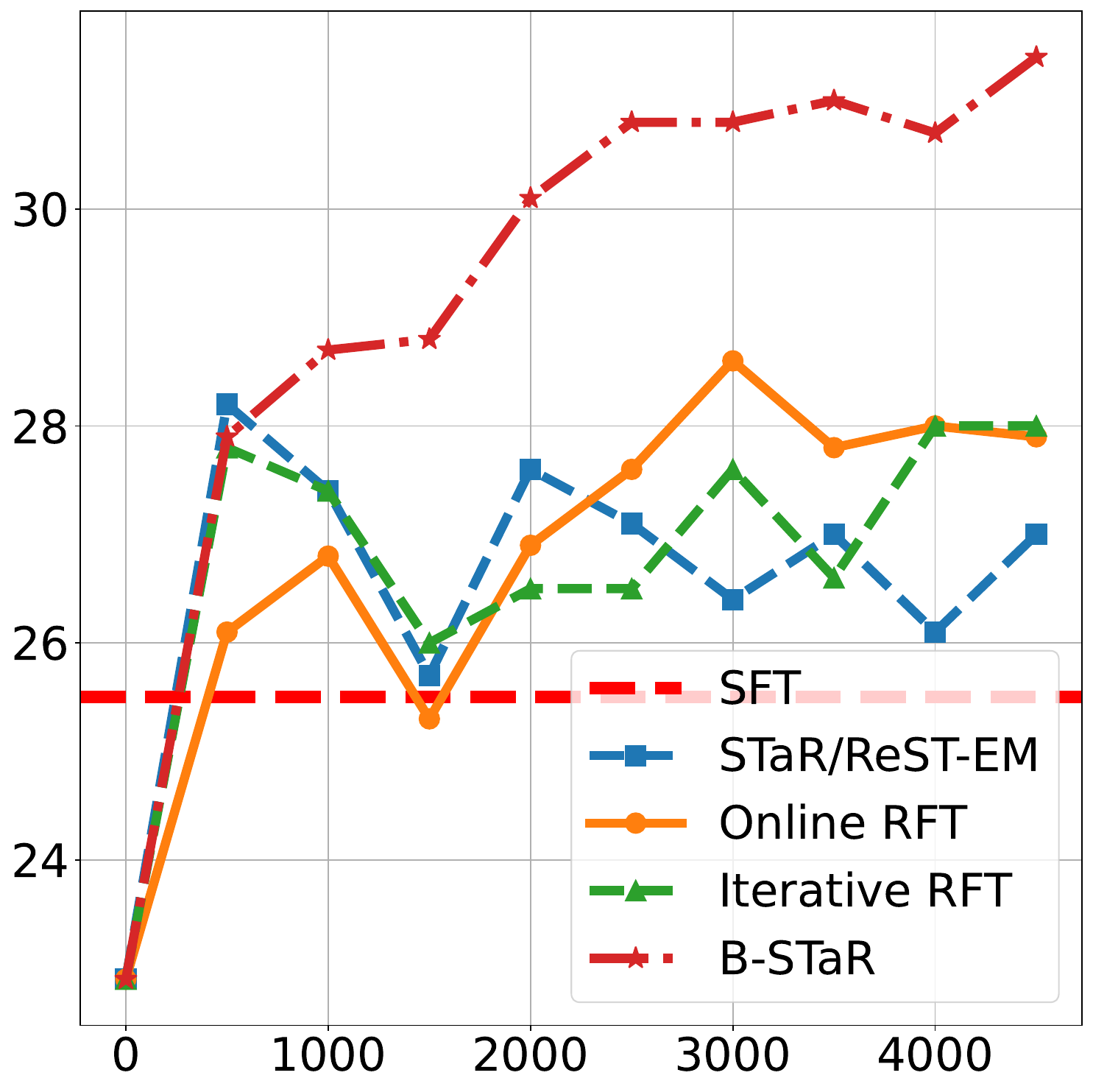}
    }
    \vspace{-5pt}
    \caption{Pass@K-S over training steps on GSM8K, MATH and APPS. For ARC-Challenge, due to the limited response space of multiple-choice questions, Pass@K and Pass@K-S metrics (where $K > 1$) provide no additional insights and are therefore excluded.}
    \label{fig:main_pass@k-s}
    \vspace{-5pt}
\end{figure*}

% \begin{table*}[t]
% \centering
% \resizebox{0.8\textwidth}{!}{
% \begin{tabular}{lccccccccc}
% \toprule
% Step              & \textbf{500} & \textbf{1000} & \textbf{1500} & \textbf{2000} & \textbf{2500} & \textbf{3000} & \textbf{3500} & \textbf{4000} & \textbf{4500} \\ \midrule
% Temperature       & 0.5          & 0.8           & 0.9           & 1             & 1.1           & 1.1           & 0.9           & 1.1           & 1.1           \\
% Reward threshold & 0            & -0.1          & -0.1          & -0.1          & -0.1          & -0.1          & -0.1          & -0.1          & -0.1          \\
% Balance Score      & 0.470        & 0.538         & 0.589         & 0.621         & 0.646         & 0.660         & 0.673         & 0.678         & 0.679         \\ 
% \bottomrule
% \end{tabular}
% }
% %\vspace{-5pt}
% \caption{Dynamic configuration adjustments by \bs{} in mathematical problem-solving. The temperature increment and reward threshold increment are both set to 0.1. Additionally, finer-grained increments for these parameters are explored in detail in Appendix~\ref{sec:fine_config} and summarized in Table~\ref{tab:b_star_hyper_fine}.}
% \label{tab:b_star_hyper}
% \vspace{-10pt}
% \end{table*}

\begin{figure*}[t]
    \centering
    \subfigure[GSM8K]{
        \includegraphics[width=0.27\textwidth]{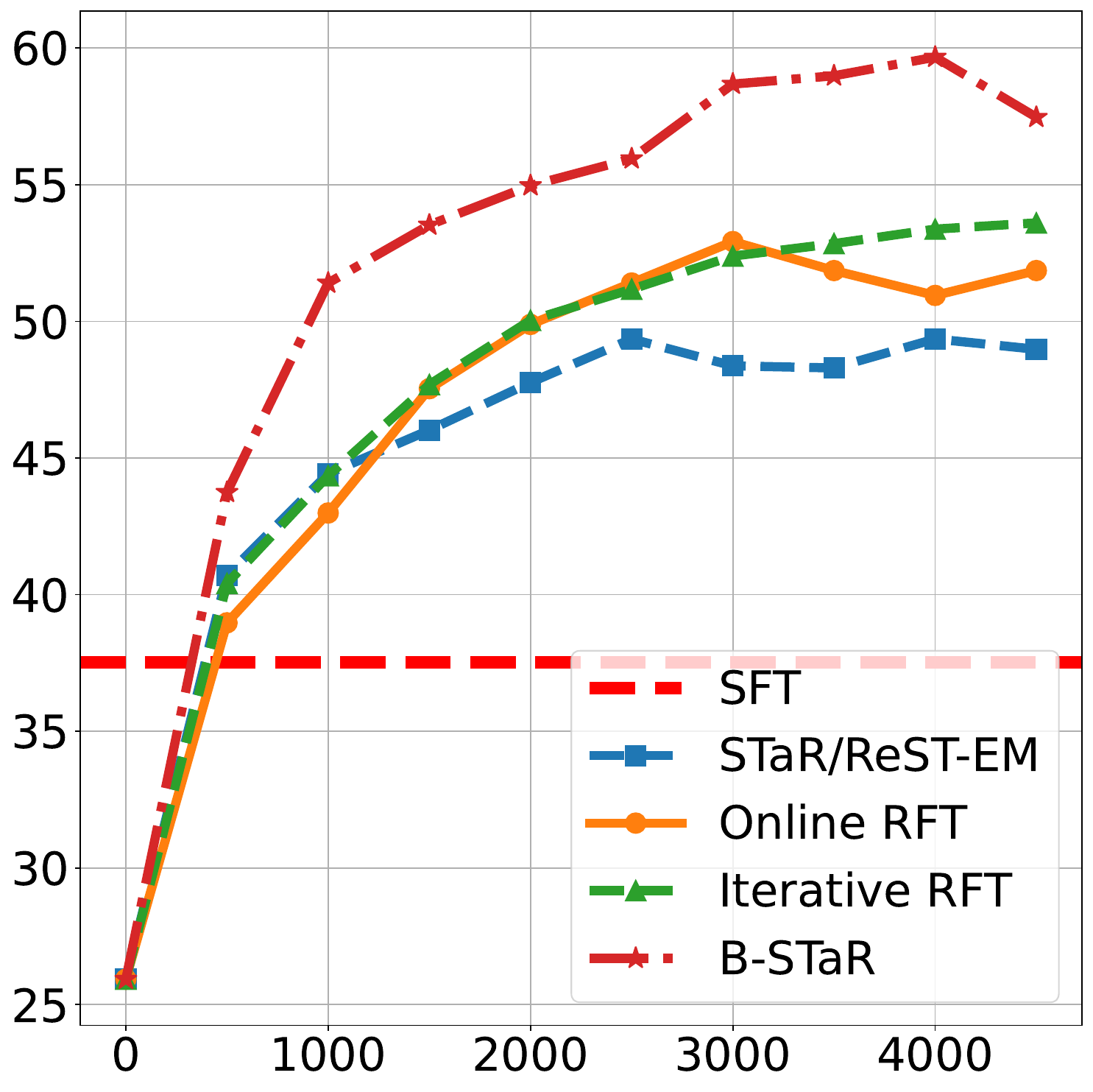}
    }
    \hspace{15pt}
    \subfigure[MATH]{
        \includegraphics[width=0.27\textwidth]{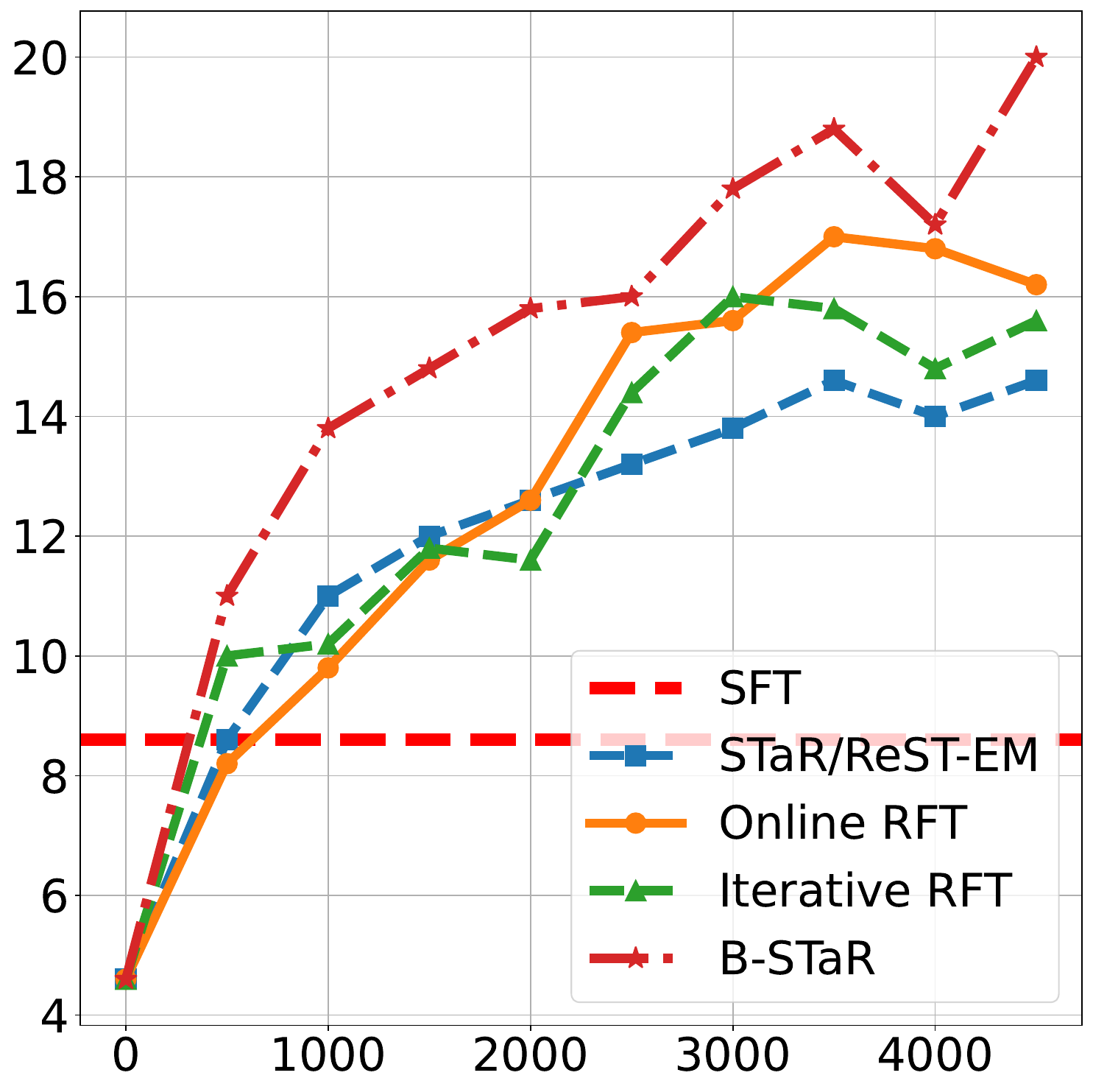}
    }
    \hspace{15pt}
    \subfigure[Balance Score]{
        \includegraphics[width=0.27\textwidth]{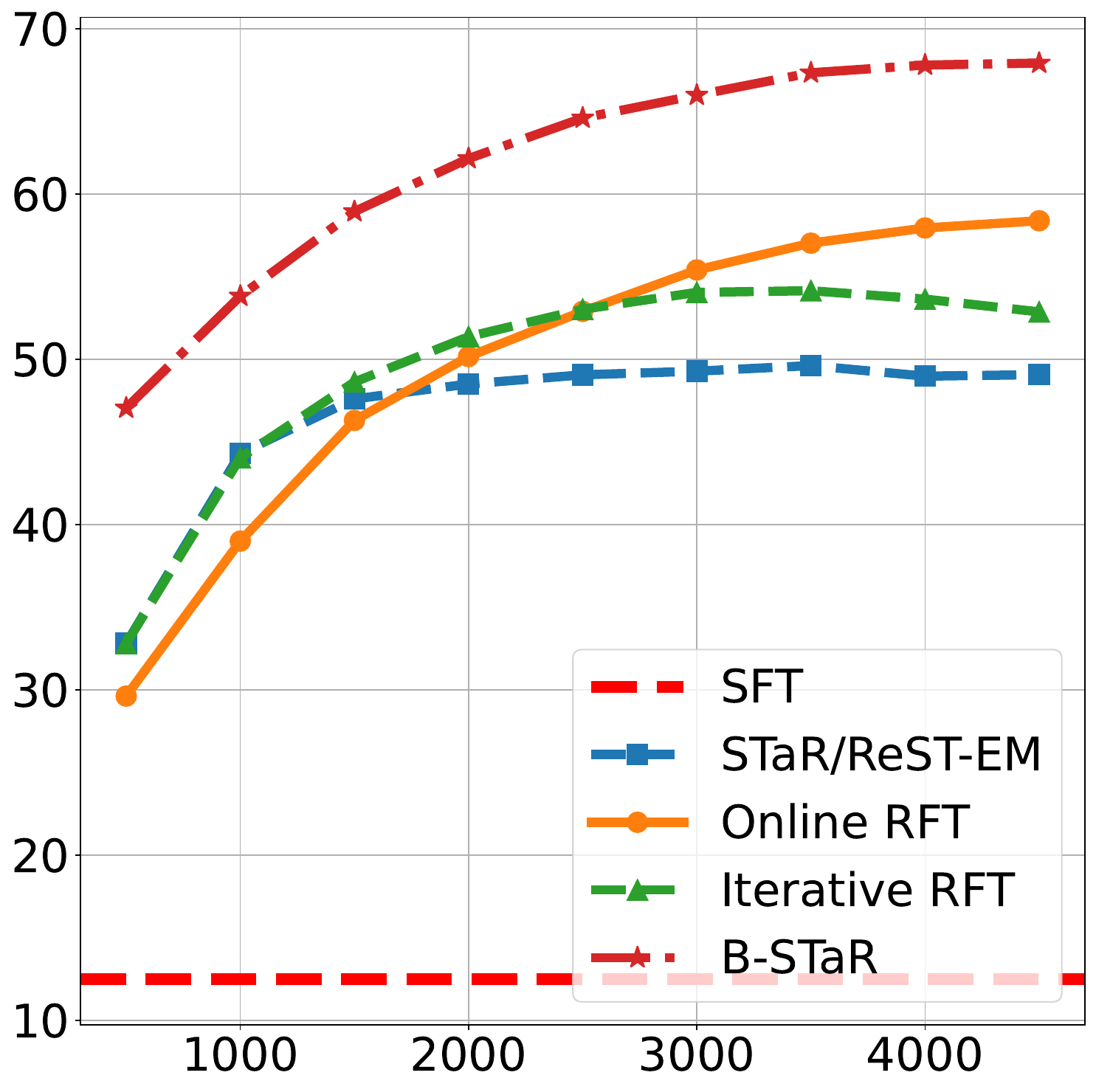}
        \label{fig:query_effect_main}
    }

    \vspace{-10pt}
    \caption{Reward@K-S over training steps on GSM8K (left), MATH (middle). For coding challenges and commonsense reasoning, we do not use a reward model, but instead use unit tests and ground truth answers respectively as binary rewards, so we do not report Reward@K-S. Balance Score over training steps (right).}
    \label{fig:main_reward@k-s}
\vspace{-5pt}
\end{figure*}

\subsection{Results}

\label{sec:math_result}
% \jh{I don't think we report results in separation of different tasks, instead I think it is better to discuss with the following structures: 
% 1. Main results: can discuss B-STAR outperforms baselines on all 3 tasks, then talk about figure 5, 6 on exploration also improving
% 2. How does the configurations change? This is Table 2 results.
% Basically we do not need to change much on the contents here, just to re-organize them
% }

% \yh{Reorganized version. If ok, plz comment out the original text. Begin below}

\textbf{Main Results.}  Table \ref{tab:b_star_main_result} provides a comprehensive comparison of \bs{} with various self-improvement methods, including Rest-EM, Iterative RFT, online RFT, and their reward model variants in mathematical problem-solving, coding challenges and commonsense reasoning. The results show that B-STAR consistently achieves higher Pass@1 scores across all scenarios, highlighting its ability to effectively steer the model toward correct responses via greedy decoding. Moreover, \bs{} demonstrates significantly better Pass@K-S values, reflecting an enhanced exploration capacity that facilitates the generation of a wider range of high-quality responses. Notably, online RFT outperforms predominantly offline methods like Rest-EM, illustrating that dynamic, on-policy approaches strike a more effective balance between learning efficiency and performance gains.

As shown in Figure~\ref{fig:main_pass@1}, all the self-improvement methods exhibit significant performance improvement after the first iteration. However, as the number of iterations increases, the growth trends of other baseline methods slows down and eventually stagnate. In contrast, \bs{} maintains a substantial growth rate and consistently outperforms other baselines. This suggests that balancing exploration and exploitation is crucial for achieving stable and efficient self-improvement.

 Moreover, in Figures~\ref{fig:main_pass@k-s} and~\ref{fig:main_reward@k-s}, we present the dynamic evolution of \bs{} and Online RFT throughout the self-improvement process in mathematical problem-solving. Figure~\ref{fig:query_effect_main} highlights \bs{}'s effectiveness in balancing exploration and exploitation-related configurations, resulting in a markedly higher balance score. This optimal balance drives the consistently higher Pass@K-S and Reward@K-S scores observed in Figures~\ref{fig:main_pass@k-s} and~\ref{fig:main_reward@k-s} across both datasets. These results suggest that \bs{} not only encourages the model to generate a diverse range of accurate responses but also efficiently incorporates feedback from the reward model. 

\begin{table*}[!t]
\centering
\resizebox{0.85\textwidth}{!}{
\begin{tabular}{lccccccccc}
\toprule
Step              & \textbf{500} & \textbf{1000} & \textbf{1500} & \textbf{2000} & \textbf{2500} & \textbf{3000} & \textbf{3500} & \textbf{4000} & \textbf{4500} \\ \midrule
Temperature       & 0.5          & 0.8           & 0.9           & 1             & 1.1           & 1.1           & 0.9           & 1.1           & 1.1           \\
Reward threshold & 0            & -0.1          & -0.1          & -0.1          & -0.1          & -0.1          & -0.1          & -0.1          & -0.1          \\
Balance Score      & 0.470        & 0.538         & 0.589         & 0.621         & 0.646         & 0.660         & 0.673         & 0.678         & 0.679         \\ 
\bottomrule
\end{tabular}
}
%\vspace{-5pt}
\caption{Dynamic configuration adjustments by \bs{} in mathematical problem-solving. The temperature increment and reward threshold increment are both set to 0.1. Additionally, finer-grained increments for these parameters are explored in detail in Appendix~\ref{sec:fine_config} and summarized in Table~\ref{tab:b_star_hyper_fine}.}
\label{tab:b_star_hyper}
\vspace{-10pt}
\end{table*}

 \textbf{Dynamic Configuration Adjustments by \bs{}.} Table~\ref{tab:b_star_hyper} presents the configurations automatically adjusted by \bs{} at various training stages, along with the resulting balance scores. Early in training, \bs{} employs a lower sampling temperature, typically around 0.5, which gradually increases as training advances. This gradual increase promotes cautious exploration during the initial phases, allowing the model to stabilize before expanding to broader sampling in later stages. Concurrently, \bs{} enforces stricter reward thresholds at the start, relaxing them as the model becomes more proficient. These strategies, as discussed in Section~\ref{sec:config_query_effect}, enable \bs{} to maintain an effective balance between exploration and exploitation throughout the training process.

\subsection{Ablation Study and Additional Results}

\begin{table*}[t]
\centering
\resizebox{0.65\textwidth}{!}{
\begin{tabular}{lcc}
\toprule
                              \multicolumn{1}{c}{Methods}           & \multicolumn{1}{l}{\textbf{GSM 8K}} & \multicolumn{1}{l}{\textbf{MATH}}    \\ \midrule
Online RFT                                       & 46.8                                & 23.2                                 \\
\bs{}  (Temperature Adjustment Only)                 & 53.1                                & 25.0                                 \\
\bs{}  (Reward Threshold Adjustment Only)               & 49.1                                & 24.6                                 \\ \rowcolor{gray!25}
\bs{} (Temperature + Reward Threshold) & \textbf{53.8}                       & {\textbf{27.8}} \\ \bottomrule
\end{tabular}
}
%\vspace{-5pt}
\caption{Ablation study on dynamic adjustment in mathematical problem-solving, including temperature adjustment only  and reward threshold adjustment only.}
% \vspace{-5pt}
\label{tab:ablation_configuration}
% \vspace{-5pt}
\end{table*}

\begin{table*}[t]
\centering
\resizebox{0.60\textwidth}{!}{
\begin{tabular}{lcccc}
\toprule
\multicolumn{1}{c}{Methods} & GSM8K & MATH & APPS & ARC-C \\ \midrule
SFT                         & 49.4   & 18.8 & 15.6 & 78.8  \\
Rest-EM (w/RM)              & 60.2   & 28.2 & 16.4 & 85.5  \\
Iterative RFT (w/RM)        & 55.3   & 27.2 & 17.1 & 85.2  \\
Online RFT (w/RM)           & 59.7   & 27.8 & 16.9 & 85.2  \\
\rowcolor{gray!25}
\bs{}                      & \textbf{61.6}   & \textbf{29.2} & \textbf{18.1} & \textbf{86.3}  \\ \bottomrule
\end{tabular}
}
% \vspace{-5pt}
\caption{A comparison of self-improvement methods trained on Llama-3.1-8B across MATH, GSM8K, APPS, and ARC-Challenge, showing the highest Pass@1 results. For ARC-Challenge, we start from Llama-3.1-8B-Instruct and omit the SFT stage due to the absence of CoT data for this dataset.}
\vspace{-10pt}
\label{tab:b_star_llama3.1b}
\vspace{-7pt}
\end{table*}

\paragraph{Ablation on Dynamic Adjustments.}
Table \ref{tab:b_star_main_result} and Table \ref{tab:b_star_hyper} illustrate the impact of dynamically adjusting both the temperature and reward threshold on balance score and overall performance. To emphasize the importance of each configuration, we perform experiments where one configuration remains at the default settings of online RFT while the other adjusts dynamically. As shown in Table \ref{tab:ablation_configuration}, dynamic adjustments to both configurations are critical, with performance significantly deteriorating when either configuration is fixed individually.
To demonstrate that \bs{}'s improvements stem from dynamic configuration adjustments rather than from suboptimal configuration settings, we compare \bs{} with online RFT using various fixed configuration combinations in the Appendix \ref{sec:fix_config}. Even the best fixed configurations from grid search underperform compared to B-STaR, underscoring the crucial role of dynamic adjustments.
% \jh{In the appendix we have hyperparameter search results but it is never cited in the main paper right? I think we may briefly mention it in this paragraph because it is also an ablation for dynamic adjustment (generally for every appendix section we should cite it somewhere in the main body)}

\paragraph{Generalizing to More Powerful Models.}

% \begin{table*}[t]
% \centering
% \resizebox{0.60\textwidth}{!}{
% \begin{tabular}{lcccc}
% \toprule
% \multicolumn{1}{c}{Methods} & GSM8K & MATH & APPS & ARC-C \\ \midrule
% SFT                         & 49.4   & 18.8 & 15.6 & 78.8  \\
% Rest-EM (w/RM)              & 60.2   & 28.2 & 16.4 & 85.5  \\
% Iterative RFT (w/RM)        & 55.3   & 27.2 & 17.1 & 85.2  \\
% Online RFT (w/RM)           & 59.7   & 27.8 & 16.9 & 85.2  \\
% \rowcolor{gray!25}
% \bs{}                      & \textbf{61.6}   & \textbf{29.2} & \textbf{18.1} & \textbf{86.3}  \\ \bottomrule
% \end{tabular}
% }
% % \vspace{-5pt}
% \caption{A comparison of self-improvement methods trained on Llama-3.1-8B across MATH, GSM8K, APPS, and ARC-Challenge, showing the highest Pass@1 results. For ARC-Challenge, we start from Llama-3.1-8B-Instruct and omit the SFT stage due to the absence of Chain-of-Thought (CoT) data for this dataset.}
% \label{tab:b_star_llama3.1b}
% %\vspace{-7pt}
% \end{table*}

To assess the generalization capability of \bs{} on more powerful models, we train the model using Llama-3.1-8b and evaluate its performance across MATH, GSM8K, APPS and ARC-Challenge. Due to computational resource constraints, we set the sample size for math reasoning to 48, and the sample size for APPS and ARC-Challenge to 32, while keeping all other parameters consistent with the\textsection\ref{sec:math_result}. The results in Table \ref{tab:b_star_llama3.1b} show that \bs{} successfully generalizes to more powerful models.

\section{Discussion}
In this paper, 
% \jhc{we conduct a comprehensive investigation into the intrinsic mechanisms of self-improvement, identifying two critical factors: (1) the model's ability to explore and generate a diverse range of high-quality responses (exploration), and (2) the reliability of external rewards for distinguishing optimal responses (exploitation). Through quantitative experiments, we track the evolution of these capabilities throughout the self-improvement process. Our findings indicate that a balance between these factors is required. To this end,
% % the model's exploration capacity declines rapidly as training progresses, while changes in the policy model's distribution diminish the effectiveness of exploiting external rewards. 
% we propose \bs{}, a method that dynamically adjusts configurations during the self-improvement process to maintain a balance between exploration and exploitation. Our experiments in mathematical reasoning, coding challenges and commonsense reasoning demonstrate that \bs{} significantly improves performance.}{
we reveal the importance of the balance between exploration and exploitation in self-improving training, and we propose \bs{}, a novel algorithm to strike a better balance and achieve superior performance on reasoning and coding tasks. While we adopt a simple method that manipulates exploration and exploitation through hyperparameter configurations, we expect more flexible control of these two factors can be achieved in the future to strike a better balance, such as advanced decoding approaches to directly control the exploration of the generated data, and reward model update to improve exploitation.

% \textbf{Limitations} Our research has several limitations. First, we focus solely on smaller LLMs, such as Mistral 7B and Llama 3 8B. Future studies should explore how exploration and exploitation capabilities evolve during the self-improvement process in larger LLMs. Second, our self-improvement experiments begin with a model that undergoes supervised fine-tuning (SFT) for only one epoch. Investigating the impact of the initial model—whether a pre-trained version or one with more extensive SFT—on exploration and exploitation is valuable. Lastly, all experiments rely on labeled queries; future research should examine the role of unlabeled queries in the self-improvement process.

% \clearpage
\bibliography{iclr2025_conference}
\bibliographystyle{iclr2025_conference}

\newpage
\appendix
\section{Experiments for the Case Study}

\subsection{Experiment Setup}

\label{sec:casestudy_setup}
\textbf{Datasets} We use the MATH dataset for training and validate the model's mathematical reasoning ability using test sets from both the MATH~\citep{hendrycks2021measuring} and GSM 8K~\citep{cobbe2021training} datasets. For the MATH dataset, we follow previous settings~\citep{lightman2023let,wang2024math,sun2024easy} by using a subset of 500 representative problems (MATH500) as our test data. We uniformly sample an additional 500 problems for validation and use the remaining 4,000 problems from the MATH test set along with the original 7,500 training problems as our training data.

\textbf{Implementation details}
For SFT, we use Mistral-7B~\citep{jiang2023mistral} as the base model with a learning rate of 5e-6, a batch size of 128, and train for 3 epochs. After the first epoch, the model (denoted as $P_{0}$) is used as the starting point for self-improvement. We then proceed with 9 iterations, where each iteration consists of 500 training steps with a batch size of 128. At the beginning of each iteration, we sample 32 candidate responses for each query, using a temperature of 1.0.

For the Process Reward Model (PRM), we automatically generate process annotations following the MATH-Shepherd approach~\citep{wang2024math}. Using the SFT model trained for 1 epoch, we sample 15 responses for each query in the training set. The SFT model trained for 3 epochs serves as the completer, decoding 8 solutions per step to annotate the sampled data. This process results in approximately 270 K process reward annotations. We train the reward model using the Mistral-7B base, with a learning rate of 2e-6, for 2 epochs. During rewarding, we use the lowest step score in the solution as the PRM Reward, normalize it to a range of [-1, 1] and combine it with a sparse reward to form the final reward score. We set the reward threshold to 0.0, selecting only those responses with final reward scores exceeding this threshold.
%This combined reward is then used to select the top 8 solutions.

\subsection{Results on GSM8K}
\begin{figure*}[!t]
    \centering
    \subfigure[Pass@1 on GSM8K]{
        \includegraphics[width=0.23\textwidth]{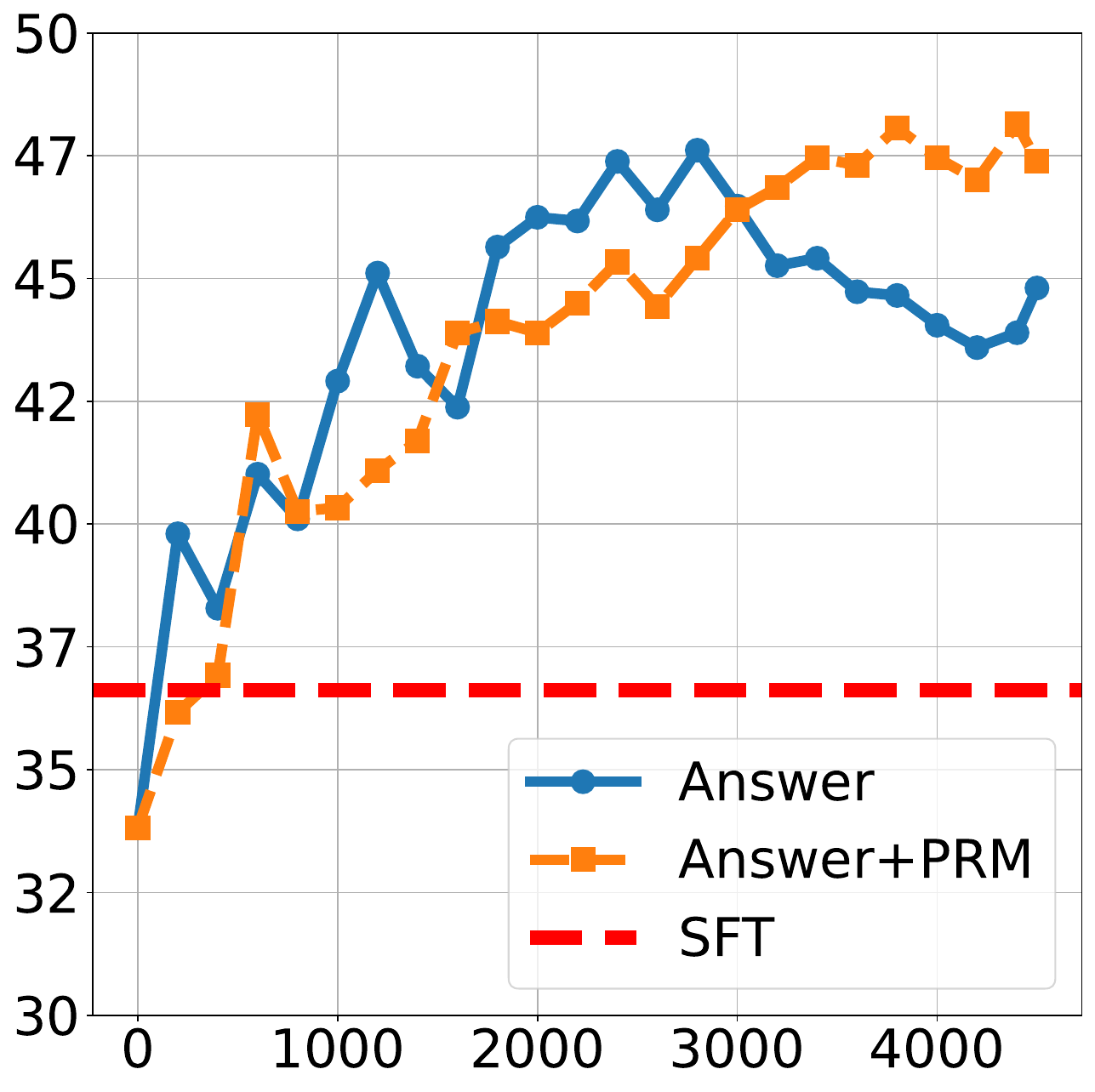}
    }
    % \subfigure[Pass@1 on MATH]{
    %     \includegraphics[width=0.23\textwidth]{fig/math_pass@1_math_only_v1.pdf}
    % }
    \subfigure[Diversity on GSM8K]{
        \includegraphics[width=0.23\textwidth]{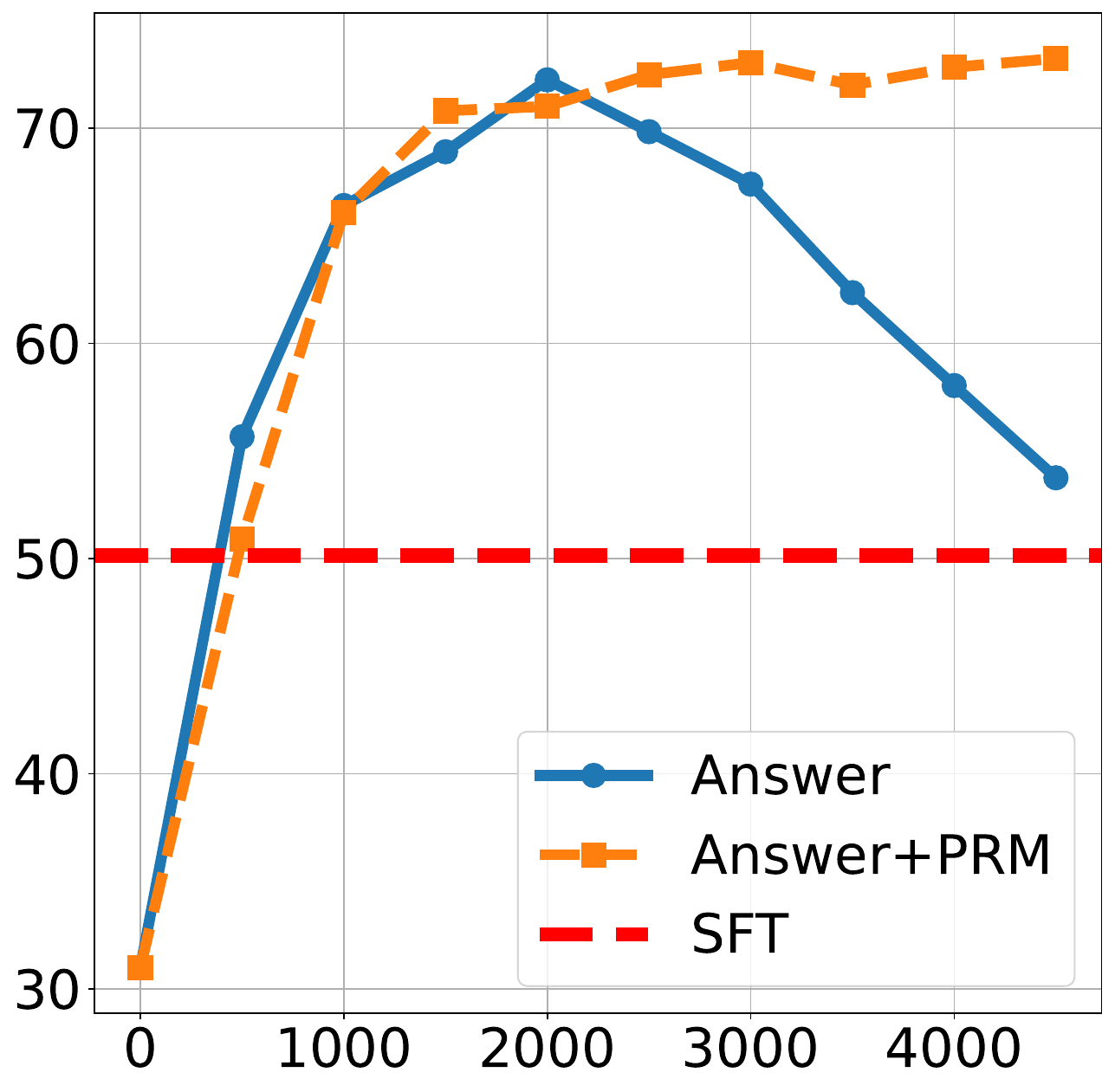}
    }
    % \subfigure[Diversity on MATH]{
    %     \includegraphics[width=0.23\textwidth]{fig/math_diversity_math_only_v1.pdf}
    % }
    % \subfigure[Pass@K-S on MATH]{
    %     \includegraphics[width=0.23\textwidth]{fig/math_pass@k-s_math_only_v1.pdf}
    % }  
    \subfigure[Pass@K-S on GSM8K]{
        \includegraphics[width=0.23\textwidth]{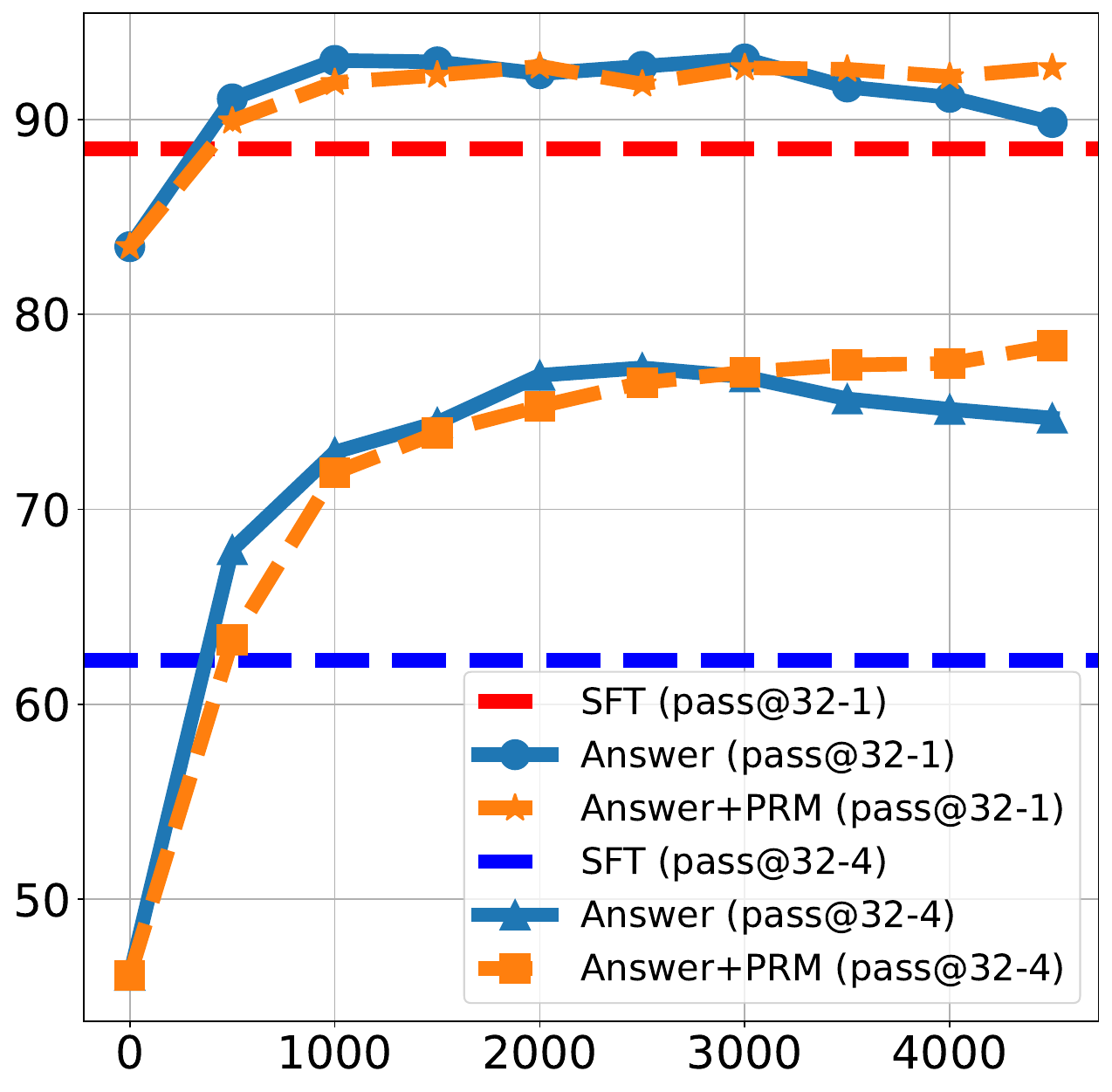}
    }
    \subfigure[Reward@K-S GSM8K]{
        \includegraphics[width=0.23\textwidth]{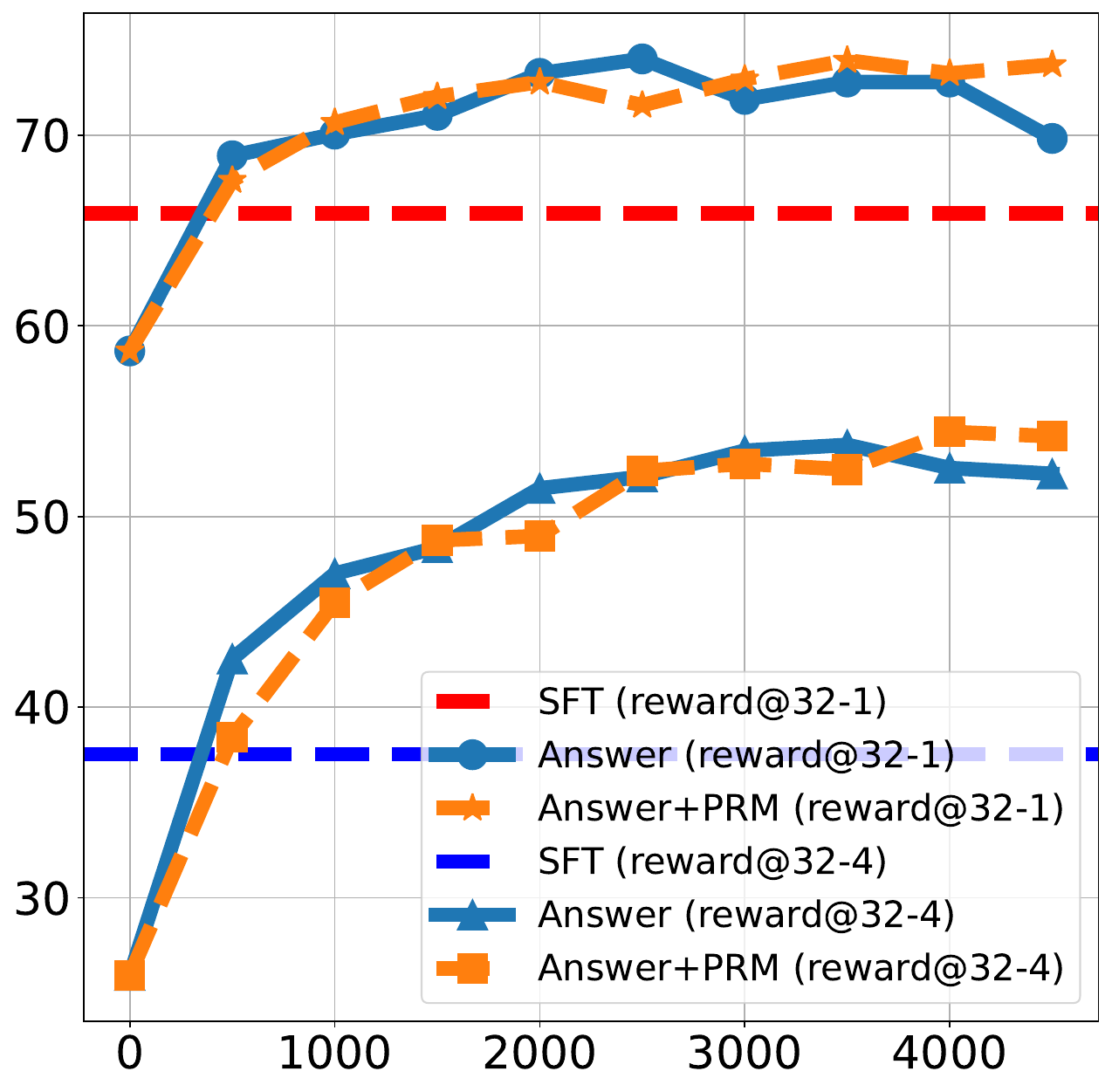}
    }
    % \subfigure[Reward@K-S MATH]{
    %     \includegraphics[width=0.23\textwidth]{fig/math_reward@k-s_math_only_v3.pdf}
    % }
    % \hspace{-5pt}
    \caption{Pass@1, Diversity, Pass@K-S and Reward@K-S over training steps on GSM8K. SFT refers to direct fine-tuning using the original dataset, "Answer" indicates matching against the ground-truth final answer, and "Answer + PRM" combines ground-truth final answer matching with PRM reward.}
    \label{fig:pass@1&diversity_gsm8k}
    % \vspace{-10pt}
\end{figure*}

In Figure \ref{fig:pass@1&diversity_gsm8k}, we present the changes in Pass@1, Diversity, Pass@K-S, and Reward@K-S metrics on GSM8K. Our results indicate that incorporating PRMs leads to additional improvements on GSM8K. However, these improvements are less pronounced on MATH. We hypothesize that this is due to the difficulty of MATH problems, which may exceed the discriminative capabilities of the 7B reward model.

\label{sec:results_on_gsm8k}

\begin{table*}[t]
\centering
\resizebox{0.85\textwidth}{!}{
\begin{tabular}{llllllllll}
\toprule
\textbf{Step}     & \textbf{500} & \textbf{1000} & \textbf{1500} & \textbf{2000} & \textbf{2500} & \textbf{3000} & \textbf{3500} & \textbf{4000} & \textbf{4500} \\ 
\midrule
Temperature       & 0.65         & 0.75          & 1.05          & 0.95          & 1.05          & 0.85          & 1.05          & 1.15          & 1.05          \\
Reward Thresholds & -0.02        & -0.04         & -0.09         & -0.09         & -0.14         & -0.14         & -0.14         & -0.15         & -0.06         \\
Balance Score      & 0.500        & 0.557         & 0.591         & 0.626         & 0.652         & 0.665         & 0.679         & 0.682         & 0.684         \\ 
\bottomrule
\end{tabular}
}
% \vspace{-5pt}
\caption{Finer-grained dynamic configuration adjustments by \bs{} in mathematical problem-solving.}
\label{tab:b_star_hyper_fine}
\vspace{-10pt}
\end{table*}

%\section{\bs{} Algorithm}

\begin{algorithm}[!t]
  \footnotesize
  %\scriptsize
  \caption{\bs{}}
  \label{alg_bstar}
  \begin{algorithmic}[1]
    \REQUIRE \# Iterations $I$, initial policy $P_0$, reward model $RM$, dataset $\mathcal{D}$, temperature $t \in \mathcal{T}$, sample size $k$, threshold $\tau \in \Theta$, optimizer $\mathcal{O}_0$, scheduler $\mathcal{L}_0$.
    \ENSURE Final updated model $P_I$.

    \FOR{$i=1$ to $I$}
      \STATE \textcolor{purple}{// Step 1: Balance Exploration \& Exploitation}
      \STATE \textcolor{purple}{// $\mathrm{BS}(\cdot)$: the average balance score (\textsection\ref{sec:query-effect})}
      \STATE $t_i, \tau_i = \arg\max_{t \in \mathcal{T},\tau \in \Theta}\mathrm{BS}(t, \tau)$

      \STATE \textcolor{purple}{// Step 2: Generate Candidates}
      \STATE Draw $M$ queries $\{x_j\}_{j=1}^M \subseteq \mathcal{D}$ 
      \FOR{$j=1$ to $M$} 
        \FOR{$m=1$ to $k$} 
          \STATE $y_{j,m} \sim P_{i-1}(\cdot \mid x_j; t_i)$ 
        \ENDFOR 
      \ENDFOR 

      \STATE \textcolor{purple}{// Step 3: Evaluate via Reward Model}
      \STATE $\mathcal{D}_i=\{(x_j,y_{j,m})\}; \quad r_{j,m}=RM(x_j,y_{j,m})$
      \STATE $\mathcal{D}_i'=\{(x_j,y_{j,m}) \mid r_{j,m}>\tau_i\}$

      \STATE \textcolor{purple}{// Step 4: Improve Policy Model}
      \STATE $P_i=\text{Update}(P_{i-1},\mathcal{D}_i',\mathcal{O}_{i-1},\mathcal{L}_{i-1})$

      \STATE \textcolor{purple}{// Step 5: Inherit Optimizer \& Scheduler}
      \STATE $\mathcal{O}_i \leftarrow \mathcal{O}_{i-1}, \; \mathcal{L}_i \leftarrow \mathcal{L}_{i-1}$

    \ENDFOR
  \end{algorithmic}
  % \vspace{-5pt}
\end{algorithm}

\section{Experiment Setup for Main Experiments}
\label{sec:main_setup}

\subsection{Baseline}
\label{sec:baseline_method}
Our evaluation compares several baseline methods: STaR/ResT-EM ~\citep{zelikman2022star,singh2023beyond}, Iterative RFT, and Online RFT~\citep{shao2024deepseekmath}. The STaR/ResT-EM approach involves multiple iterations, where each iteration samples from the latest policy model but resets and retrains the model from scratch. In contrast, Iterative RFT builds on each previous iteration by inheriting the checkpoint and resuming training from that point. Online RFT goes a step further by not only inheriting the checkpoint but also the optimizer state and learning rate schedule, ensuring a smoother transition across iterations.

\subsection{mathematical problem-solving}
For mathematical problem-solving, we largely maintain the experimental setup from \textsection\ref{sec:case_study}. We use the Mistral-7B model as our base and conduct SFT on the MATH training dataset for three epochs. The first epoch serves as the starting point for the model's self-improvement phase. We set the number of samples per iteration ($N$) to 67,500 and feed 11,500 MATH training queries ($M$) per iteration. We set sample size to 64 for all methods. 
We vary temperature from 0.5 to 1.2 in 0.1 increments and reward threshold from -1.0 to 1.0 in 0.1 increment. Additionally, finer increments for both the temperature and reward threshold are explored in Appendix~\ref{sec:fine_config}. 
Throughout the self-improvement process, we use Pass@1, Pass@K-S, and Reward@K-S metrics to track changes in the performance and model's exploration and exploitation capabilities.

\subsection{Coding Challenges}
On coding challenges, we follow ~\citet{singh2023beyond} and adopt the APPS~\citep{hendrycks2021measuring} dataset for both training and testing. To balance the number of responses per question, we sample 5 responses per question from the original APPS training set forming a dataset with 13K examples. We use Llama-3-8B~\citep{dubey2024llama} as our base model,\footnote{We do not use Mistral-7B as in the math domain because we found it performs poorly on coding tasks.} keeping the rest of the settings consistent with those of the math domain. 
% Similarly, we start from the first SFT checkpoint to conduct our self-improvement experiments. For the baseline method, 
For baselines, we uniformly sample 32 candidate responses per query with a temperature of 0.4. For \bs{}, we explore temperatures 0.4 to 1.1 in 0.1 increment to determine the optimal configuration. We set the number of samples per iteration ($N$) to 13,500 and feed 2627 APPS training queries ($M$) per iteration. We do not apply reward models to the coding task and instead use unit tests as the binary reward, which means only the sampling temperature is automatically adjusted in \bs.

\subsection{commonsense reasoning}
For commonsense reasoning, following~\citet{pang2024iterative}, we conduct experiments on ARC-Challenge~\citep{clark2018think}, a dataset consisting of multiple-choice science questions designed to evaluate commonsense reasoning beyond mathematics and coding challenges. We start with the Mistral-7b-instruct model and omit the SFT stage due to the absence of the CoT data for this dataset. Other configurations, such as sample size and temperature, are the same as those used in the coding tasks. The ground-truth answer serves as the binary reward, and we report only Pass@1 results for the ARC-Challenge dataset. Given the constrained response space inherent to multiple-choice questions, Pass@K and Pass@K-S metrics (where $K>1$) yield no additional insights and are therefore excluded.

\section{Details of Self-Improvement}
\label{sec:concept_self}

\textbf{Fixed Reward Function} \\ A fixed reward function $r(x,y)$ is a predefined, static function that does not adapt based on the training process or model parameters. For instance, binary feedback (e.g., whether a math problem is solved correctly or a unit test passes) is an example of a fixed reward function. \\ The fixed reward function can be represented as: 
\begin{equation}
r(x, y) = 
\begin{cases}
1, & \text{if } y \text{ satisfies a predefined condition (e.g., test passes)}, \\
0, & \text{otherwise.}
\end{cases}
\end{equation}

\textbf{Reward Model} \\ A trained reward function \( r(x, y; \phi) \) is parameterized by \( \phi \) and adapts based on the training process. These reward functions (such as output-based reward models (ORMs) or process-based reward models (PRMs)) are learned from data, where the model learns to assign continuous scores based on supervision signals. \\ The trained reward model can be represented as:
\begin{equation}
r(x, y; \phi) = f(x, y; \phi)
\end{equation}
where \( f(x, y; \phi) \) is a learned function (e.g., a neural network) that maps the input \( x \) and the response \( y \) to a continuous reward score.

\textbf{Rejection Sampling Fine-tuning} \\Rejection Sampling Fine-tuning (RFT) first samples multiple outputs from the supervised fine-tuned LLMs for each query and then trains LLMs on the sampled responses with the correct answer. Formally, the objective of RFT is to maximize the following objectives:
\begin{equation}
\mathcal{L}_{\text{RFT}}(\theta) = \mathbb{E}_{x \sim \mathcal{D}} \left [ 
 \mathbb{I}(y) \log p(y \mid x; \theta)  
 \right ] 
\end{equation}
The indicator function $\mathbb{I}(y)$ is defined as:
\begin{equation}
    \mathbb{I}(y) =
\begin{cases} 
1, & \text{if the answer of } y \text{ is correct}, \\
0, & \text{if the answer of } y \text{ is incorrect}.
\end{cases}
\end{equation}
Our evaluation compares several baseline methods: STaR/ResT-EM, Iterative RFT, and Online RFT. The STaR/ResT-EM approach involves multiple iterations, where each iteration samples from the latest policy model but resets and retrains the model from scratch. In contrast, Iterative RFT builds on each previous iteration by inheriting the checkpoint and resuming training from that point. Online RFT goes a step further by not only inheriting the checkpoint but also the optimizer state and learning rate schedule, ensuring a smoother transition across iterations. Additionally, we evaluate two variants: "without RM" (Answer) and "with RM" (Answer+PRM), as described in Section \textsection~\ref{sec:case_study}.

% \begin{figure*}[t]
%     \centering
%     \subfigure[GSM 8K]{
%         \includegraphics[width=0.45\textwidth]{fig/gsm8k_pass@1_math_only_v1_fine.pdf}
%         \label{fig:gsm8k_fine}
%     }
%     \subfigure[MATH]{
%         \includegraphics[width=0.45\textwidth]{fig/math_pass@1_math_only_v1_fine.pdf}
%         \label{fig:math_fine}
%     }
%     % \vspace{-5pt}
%     \caption{\add{Pass@1 accuracy over training steps on GSM 8K and MATH for fine-grained configuration adjustments versus coarse-grained configuration adjustments. The coarse-grained configuration adjustments are consistent with our settings in Section \ref{sec:bstar_setup}}}
%     \label{fig:fine_performance}
%     % \vspace{-10pt}
% \end{figure*}

\section{More Fine-Grained Configuration Adjustments}
\label{sec:fine_config}

In Section \ref{sec:bstar_setup}, we initially set the increments for both the temperature and reward threshold to 0.1. To explore the effects of using finer-grained increments on \bs{}, we further conduct finer-grained hyper-parameters search with the granularity of 0.05 for temperature and 0.01 for reward threshold in mathematical problem-solving setting. Table \ref{tab:b_star_hyper_fine} illustrates how \bs{} dynamically adjusts its configuration and the resulting impact on balance score. A comparison of Table \ref{tab:b_star_hyper} and Table \ref{tab:b_star_hyper_fine} reveals that finer-grained configuration adjustments introduce more dynamic changes to temperature and reward thresholds throughout the training process, resulting in significantly higher balance score.

\section{Impact of Fixed Configuration Combinations}
\label{sec:fix_config}
To confirm that the improvements achieved by \bs{} are due to its dynamic configuration adjustments rather than suboptimal configuration settings, we conduct a grid search to evaluate different configuration combinations for online RFT. The temperature values are selected from the set [0.5,0.7,0.9,1.1] and the reward thresholds are chosen from [-0.4,-0.2,0.0,0.2,0.4]. For comparison, we also include two specific configurations: the default combination from our paper, (1.0,0.0), and the parameters obtained from \bs{}'s final iteration, (1.1,-0.1).

\begin{figure*}[t]
    \centering
    \subfigure[GSM 8K]{
        \includegraphics[width=0.48\textwidth]{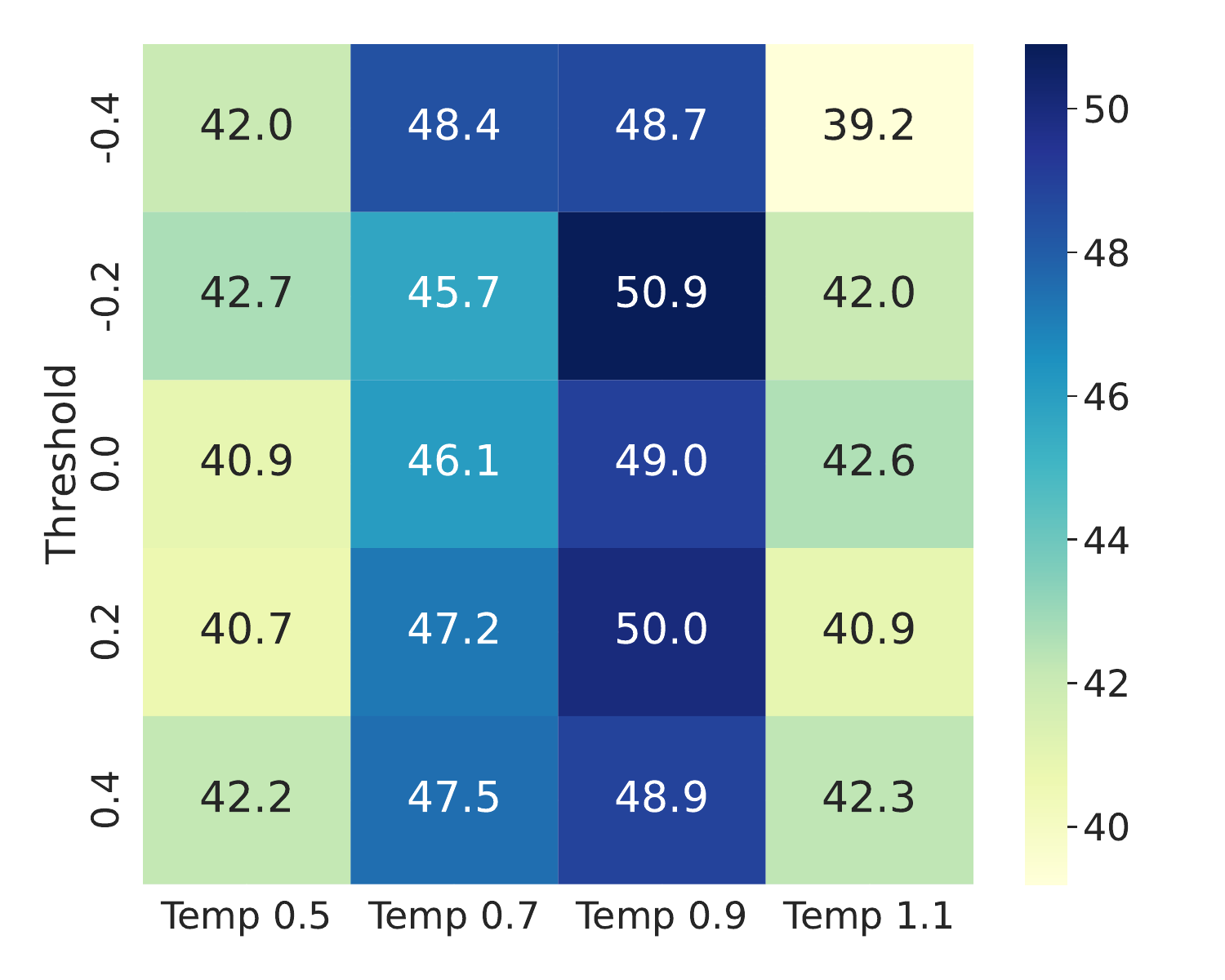}
        \label{fig:gsm8k_heat}
    }
    \subfigure[MATH]{
        \includegraphics[width=0.48\textwidth]{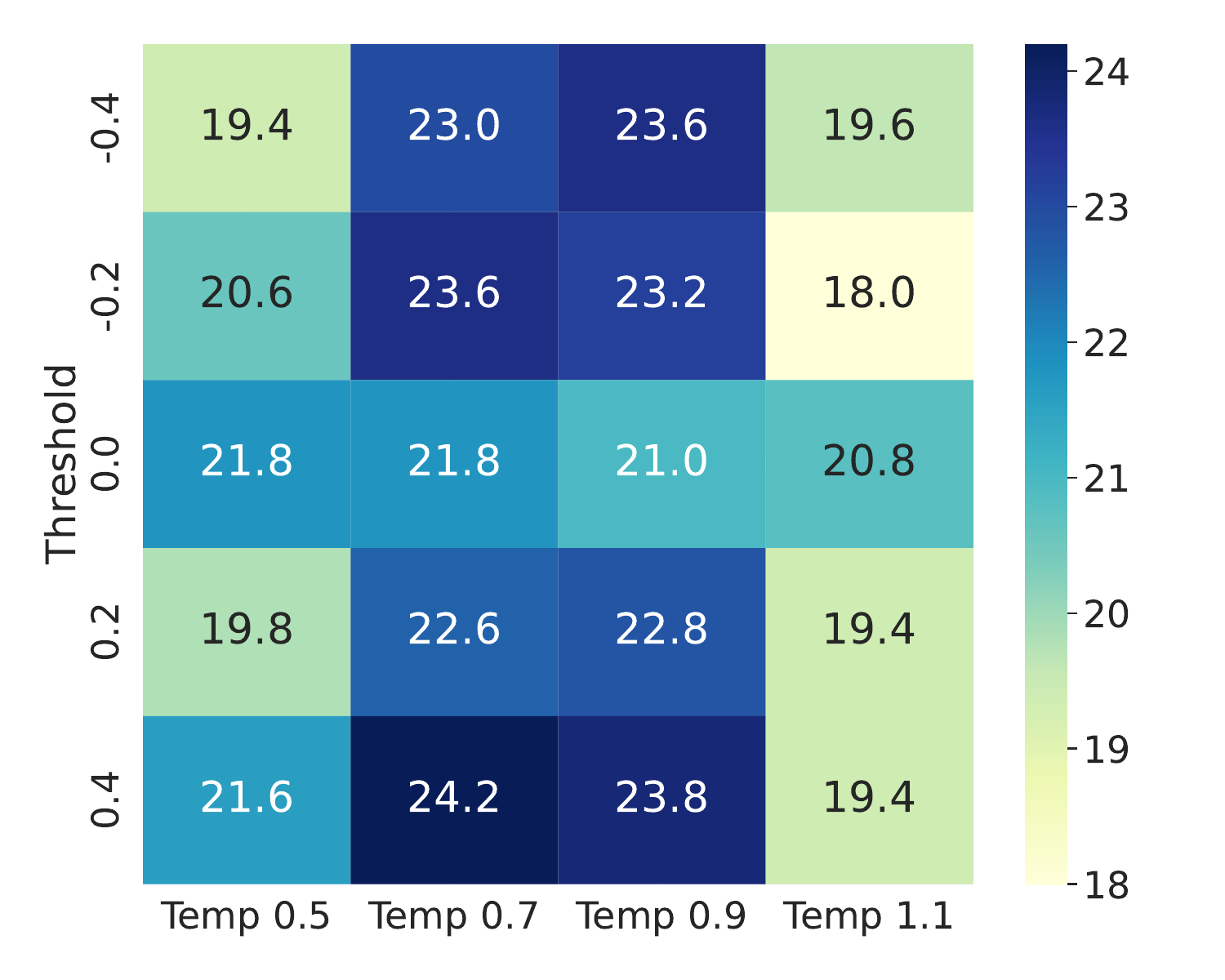}
        \label{fig:math_heat}
    }
    % \vspace{-5pt}
    \caption{Performance of online RFT using different configuration combinations, where the horizontal axis represents changes in temperature and the vertical axis represents changes in reward threshold. \bs{}'s GSM8K accuracy is 53.1\%, while its MATH accuracy is 27.8\%.}
    \label{fig:heat_map}
    % \vspace{-10pt}
\end{figure*}

Figure \ref{fig:heat_map} and Table \ref{tab:sepcific_configuration} illustrate that while grid search-based configuration combinations offer some performance improvements for online RFT, they remain less effective compared to the dynamic configuration adjustments enabled by \bs{}. This further emphasizes the critical need for dynamically balancing exploration and exploitation throughout the training process.

\begin{table*}[t]
\centering
\resizebox{0.5\textwidth}{!}{
\begin{tabular}{lcc}
\toprule
\textbf{Configuration}       & \textbf{GSM 8K} & \textbf{MATH} \\ \midrule
Temp = 1.0; Threshold = 0.0  & 46.8            & 23.2          \\
Temp = 1.1; Threshold = -0.1 & 40.4            & 18.2          \\
B-STaR                       & 53.1            & 27.8          \\ \bottomrule
\end{tabular}
}
% \vspace{-5pt}
\caption{Comparison of Online RFT using specific configurations and B-STaR Performance. This table reports the results with the stable hyperparameter combinations we found in our B-STaR experiments (Temperature = 1.1, Reward thresholds = -0.1) 
 (Table~\ref{tab:b_star_hyper}).}

\label{tab:sepcific_configuration}
% \vspace{-10pt}
\end{table*}

\section{Related Work of Dynamic Hyperparameter Adjustment}
\label{sec:related_work}

 Dynamic hyperparameter optimization addresses the shortcomings of static configurations, which cannot adapt to the evolving dynamics of machine learning. Early work by ~\citet{loshchilov2016sgdr} introduced the use of a cosine function to modulate learning rates, ensuring smoother convergence. Building on this, ~\citet{baydin2017online} proposed gradient-based methods to dynamically adjust learning rates in real-time by analyzing gradient trends, significantly accelerating convergence. Expanding the scope, ~\citet{smith2018disciplined} introduced a systematic approach to setting hyperparameters, such as learning rate, batch size, momentum, and weight decay, while highlighting their interdependence to improve training efficiency. Subsequently, ~\citet{jaderberg2017population} integrated model and hyperparameter optimization by asynchronously evolving a population of models through performance-based selection and mutation. ~\citet{jomaa2019hyp} framed hyperparameter tuning as a sequential decision-making problem, leveraging reinforcement learning to learn a policy for efficient hyperparameter selection. More recently, ~\citet{baik2020meta} adopted a meta-learning framework to dynamically generate task- and step-specific hyperparameters, improving inner-loop optimization in few-shot learning tasks. Inspired by these innovations, our approach focuses on dynamically monitoring and balancing configurations between exploration and exploitation, optimizing the synergy between current policies and reward mechanisms to drive further performance gains.

 Dynamic adjustment of hyperparameters has also proven to be crucial in pure reinforcement learning (RL) settings. For instance, ~\citet{kiran2022hyperparameter} highlights how adapting hyperparameters can significantly influence both the learning process and processing times in deep RL problems. Similarily, ~\citet{franke2020sample} proposes a population-based automated RL framework that concurrently optimizes hyperparameters and neural architectures during agent training. Furthermore, ~\citet{mohan2023autorl} examines the dynamic nature of hyperparameter landscapes in RL, offering empirical evidence that these landscapes evolve over time, varying across algorithms and environments. To enhance adaptability, ~\citet{bai2024generalized} introduces a refined framework that emphasizes granularity and flexibility in hyperparameter adjustments, incorporating a Pairwise Learning approach to provide comprehensive guidance for improving the performance of underperforming agents.

\section{Theoretical Justification for Exploration and Exploitation}
\label{appendix:just-exp}
The objective of self-improvement can be expressed in the framework of reinforcement learning as follows:
\begin{equation}
    \pi_\theta^{t+1} = \arg\max_{\pi_\theta^t}  \mathbb{E}_{x,y^* \sim \mathcal{D}, \hat{y} \sim \pi_\theta^t[\cdot|x]} \left[ R( \hat{y}, y^*) \right]
\end{equation}
where $\mathcal{D}$ represents the dataset, $x$ and $y^*$  denote the sampled input and its corresponding ground-truth answer, respectively, and $\hat{y}$ is the sampled response. Here,  $\pi_\theta^t$  corresponds to the language model in the $t$-th iteration. $R$ is the reward function. According to the policy gradient algorithm:
\begin{equation}
    \nabla_{\theta} \mathbb{E}_{x,y^* \sim \mathcal{D}, \hat{y} \sim \pi_\theta^t[\cdot|x]} \left[ R( \hat{y}, y^*) \right] = \mathbb{E}_{x,y^* \sim \mathcal{D}, \hat{y} \sim \pi_\theta^t[\cdot|x]}\nabla_{\theta}R(\hat{y}, y^*)\log \pi_\theta^t[\hat{y}|x]
\end{equation}
When $R(\hat{y}, y^*)$ is binary, the above equation turns to be simple data selection and supervised training loss that is exactly what we are doing. Thus, self-improvement can be viewed as a  form of reinforcement learning, where maintaining a balance between exploration and exploitation is crucial  and has been studied for years ~\citep{csimcsek2006intrinsic, sutton2018reinforcement, weng2018bandit, enwiki:1247645791}. Conceptually in classic RL, exploration involves exploring the environment (analogous to reasoning tasks in our paper) by trying random actions (corresponding to sampling multiple candidates in our work) to gather more information about the environment. Exploitation, on the other hand, involves utilizing the known information to maximize the reward (similar to how we use the reward function to select data samples). Insufficient exploration may cause the training process to stagnate, while insufficient exploitation can lead to instability and large variance during training.

\end{document}